\documentclass[twoside,11pt]{article}

\usepackage{jmlr2e}
\usepackage{amsmath}
\usepackage{cancel}
\usepackage{xfrac}
\usepackage{multirow}

\numberwithin{table}{section}
\numberwithin{figure}{section}

\newcommand{\mref}[1]{(\ref{#1})}
\newcommand{\E}{\operatorname{E}}
\renewcommand{\H}{\operatorname{H}}
\newcommand{\COR}{\operatorname{COR}}
\newcommand{\AR}{A_{\mathrm{REF}}}
\newcommand{\G}{\mathcal{G}}
\newcommand{\GR}{\mathcal{G}_{\mathrm{REF}}}
\newcommand{\Gp}{\mathcal{G}^{+}}
\newcommand{\Gm}{\mathcal{G}^{-}}
\newcommand{\D}{\mathcal{D}}
\newcommand{\B}{\mathcal{B}}
\newcommand{\X}{\mathbf{X}}
\newcommand{\PX}{\Pi_{X}}
\newcommand{\PXi}{\Pi_{X_i}}
\newcommand{\XP}{X \given \PX}
\newcommand{\XPi}{X_i \given \PXi}
\newcommand{\TT}{\Theta_{X_i} \given \PXi}
\newcommand{\BIC}{\operatorname{BIC}}
\newcommand{\BD}{\operatorname{BD}}
\newcommand{\BDeu}{\operatorname{BDeu}}
\newcommand{\BDs}{\operatorname{BDs}}

\newcommand{\given}{\operatorname{|}}
\newcommand{\Prob}{\operatorname{P}}
\newcommand{\tq}{\tilde{q}_i}
\newcommand{\ta}{\tilde{\alpha}}
\newcommand{\aijk}{\alpha_{ijk}}
\newcommand{\piijk}{\pi_{ij \given k}}
\newcommand{\pijk}{p_{ij \given k}}
\newcommand{\tai}{\tilde{\alpha}_{i}}
\newcommand{\asi}{\alpha^*_{i}}
\newcommand{\bsi}{\beta^*_{i}}
\newcommand{\jpo}{j: n_{ij} > 0}
\newcommand{\e}[1]{\times 10^{#1}}
\newcommand{\pra}{\overrightarrow{p_{ij}}}
\newcommand{\pla}{\overleftarrow{p_{ij}}}
\newcommand{\pri}{\mathring{p_{ij}}}
\newcommand{\fh}{\sfrac{1}{2}}

\newcommand{\dEP}[1]{d_\mathrm{EP}^{(#1)}}

\def\bbbone{{\mathchoice {\rm 1\mskip-4mu l} {\rm 1\mskip-4mu l}
            {\rm 1\mskip-4.5mu l} {\rm 1\mskip-5mu l}}}

\ShortHeadings{Beyond Uniform Priors in BN Structure Learning}{Scutari}
\firstpageno{1}

\begin{document}

\title{Beyond Uniform Priors in Bayesian Network Structure Learning}

\author{\name Marco Scutari \email scutari@stats.ox.ac.uk \\
       \addr Department of Statistics \\
       University of Oxford \\ 
       Oxford, United Kingdom}

\editor{\ldots}

\maketitle

\begin{abstract}%
  Bayesian network structure learning is often performed in a Bayesian setting,
  evaluating candidate structures using their posterior probabilities for a
  given data set. Score-based algorithms then use those posterior probabilities
  as an objective function and return the \emph{maximum a posteriori} network as
  the learned model. For discrete Bayesian networks, the canonical choice for a
  posterior score is the Bayesian Dirichlet equivalent uniform (BDeu) marginal
  likelihood with a uniform (U) graph prior, which assumes a uniform prior both
  on the network structures and on the parameters of the networks.
  In this paper, we investigate the problems arising from these assumptions,
  focusing on those caused by small sample sizes and sparse data. We then propose
  an alternative posterior score: the Bayesian Dirichlet sparse (BDs) marginal
  likelihood with a marginal uniform (MU) graph prior. Like U+BDeu, MU+BDs does
  not require any prior information on the probabilistic structure of the data
  and can be used as a replacement noninformative score. We study its theoretical
  properties and we evaluate its performance in an extensive simulation study,
  showing that MU+BDs is both more accurate than U+BDeu in learning the
  structure of the network and competitive in predicting power, while not being
  computationally more complex to estimate.
\end{abstract}


\section{Introduction}

Bayesian networks \citep[BNs;][]{pearl,koller} are a class of probabilistic
models composed by a set of random variables $\X = \{X_1, \ldots, X_N\}$ and by
a directed acyclic graph (DAG) $\G = (\mathbf{V}, A)$ in which each node in
$\mathbf{V}$ is associated with one of the random variables in $\X$ (they are
usually referred to interchangeably). The arcs in $A$ express direct dependence
relationships between the variables in $\X$; graphical separation of two nodes
implies the conditional independence of the corresponding random variables. In
principle, there are many possible choices for the joint distribution of $\X$;
literature has focused mostly on discrete BNs \citep{heckerman}, in which both
$\X$ and the $X_i$ are multinomial random variables and the parameters of
interest are the conditional probabilities associated with each variable,
usually represented as conditional probability tables. Other possibilities
include Gaussian BNs \citep{heckerman3} and conditional linear Gaussian BNs
\citep{lauritzen}.

The task of learning a BN from data is performed in two steps in an inherently
Bayesian fashion. Consider a data set $\D$ and a BN $\B = (\G, \X)$. If we 
denote the parameters of the joint distribution of $\X$ with $\Theta$, we can
assume without loss of generality that $\Theta$ uniquely identifies $\X$ in the
family of distributions chosen to model $\D$ and write 
\begin{align}
\label{eq:lproc}
  &\underbrace{\Prob(\B \given \D) = 
  \Prob(\G, \Theta \given \D)}_{\text{learning}}& &=&
    &\underbrace{\Prob(\G \given \D)}_{\text{structure learning}}& &\cdot&
    &\underbrace{\Prob(\Theta \given \G, \D)}_{\text{parameter learning}}.
\end{align}

\emph{Structure learning} consists in finding the DAG $\G$ that encodes the
dependence structure of the data. Three general approaches to learn $\G$ from
$\D$ have been explored in the literature: constraint-based, score-based and
hybrid. Constraint-based algorithms use conditional independence tests such as
mutual information \citep{itheory} to assess the presence or absence of 
individual arcs in $\G$. Score-based algorithms are typically heuristic search
algorithms and use a goodness-of-fit score such as BIC \citep{schwarz} or the
Bayesian Dirichlet equivalent uniform (BDeu) marginal likelihood 
\citep{heckerman} to find an optimal $\G$. For the latter a uniform (U) prior
over the space of DAGs is usually assumed for simplicity. Hybrid algorithms
combine the previous two approaches, using conditional independence tests to
restrict the search space in which to perform a heuristic search for an optimal
$\G$. For some examples, see \citet{hiton1}, \citet{larranaga}, \citet{cussens}
and \citet{mmhc}. 

\emph{Parameter learning} involves the estimation of the parameters $\Theta$
given the DAG $\G$ learned in the first step.  Thanks to the Markov property
\citep{pearl}, this step is computationally efficient because if the data are
complete the \emph{global distribution} of $\X$ decomposes into
\begin{equation}
\label{eq:parents}
  \Prob(\X \given \G) = \prod_{i=1}^N \Prob(\XPi)
\end{equation}
and the \emph{local distribution} associated with each node $X_i$ depends only
on the configurations of the values of its parents $\PXi$. Note that this
decomposition does not uniquely identify a BN; different DAGs can encode the
same global distribution, thus grouping BNs into equivalence classes
\citep{chickering} characterised by the skeleton of $\G$ (its underlying 
undirected graph) and its v-structures (patterns of arcs of the type 
$X_j \rightarrow X_i \leftarrow X_k$, with no arc between $X_j$ and $X_k$). 

In the remainder of this paper we will focus on discrete BN structure learning
in a Bayesian framework. In Section \ref{sec:bdeu} we will describe the
canonical marginal likelihood used to identify \emph{maximum a posteriori} (MAP)
DAGs in score-based algorithms, BDeu, and the uniform prior U over the space of
the DAGs. We will review and discuss their underlying assumptions, fundamental
properties and known problems. In Section \ref{sec:bds} we will introduce a 
new posterior score, which we will call the \emph{Bayesian Dirichlet
sparse} (BDs) marginal likelihood with a \emph{marginal uniform} (MU) prior,
and the corresponding alternative set of assumptions. We will study its
theoretical properties and we will show that it does not suffer from the same
problems as U+BDeu when learning BNs from small and sparse samples. Based on
the results of an extensive simulation study, in Section \ref{sec:sim} we will
show that MU+BDs is preferable to U+BDeu because it is more accurate in learning
$\G$ from the data; and because the resulting BNs provide predictive power that
is at least as good as that of the BNs learned using U+BDeu. Proofs for all 
theorems are collected in Appendix \ref{app:proofs} and detailed simulation
results are reported in Appendix \ref{app:tables}.

\section{U+BDeu: A Posterior Score Arising from Uniform Priors}
\label{sec:bdeu}

Starting from \mref{eq:lproc}, we can decompose $\Prob(\G \given \D)$ into
\begin{equation*}
  \Prob(\G \given \D) \propto \Prob(\G)\Prob(\D \given \G) =
    \Prob(\G)\int \Prob(\D \given \G, \Theta) \Prob(\Theta \given \G) d\Theta
\end{equation*}
where $\Prob(\G)$ is the prior distribution over the space of the DAGs and 
$\Prob(\D \given \G)$ is the marginal likelihood of the data given $\G$ averaged
over all possible parameter sets $\Theta$. Using \mref{eq:parents} we can then
decompose $\Prob(\D \given \G)$ into one component for each node as follows:
\begin{equation}
\label{eq:structlearn}
  \Prob(\D \given \G) = \prod_{i=1}^N \Prob(\XPi) =
    \prod_{i=1}^N \left[ \int \Prob(\XPi, \Theta_{X_i})
    \Prob(\TT) d\Theta_{X_i} \right].
\end{equation}
In the case of discrete BNs, we assume  $\XPi \sim \mathit{Multinomial}(\TT)$
where the $\TT$ are the conditional probabilities
$\piijk = \Prob(X_i = k \given \PXi = j)$. We then assume a conjugate prior 
$\TT \sim \mathit{Dirichlet}(\aijk)$, $\sum_{jk} \aijk = \alpha_i > 0$ to obtain
the posterior $\mathit{Dirichlet}(\aijk + n_{ijk})$ which we use to estimate the
$\piijk$ from the counts $n_{ijk}, \sum_{ijk} n_{ijk} = n$ observed in $\D$. 
$\alpha_i$ is known as the \emph{imaginary} or \emph{equivalent sample size}
and determines how much weight is assigned to the prior in terms of the size of
an imaginary sample supporting it.

\subsection{The Bayesian Dirichlet Equivalent Uniform Score (BDeu)}

Further assuming \emph{positivity} ($\piijk > 0$), \emph{parameter independence}
($\piijk$ for different parent configurations are independent),
\emph{parameter modularity} ($\piijk$ associated with different nodes are
independent) and \emph{complete data}, \citet{heckerman} derived
a closed form expression for \mref{eq:structlearn}, known as the \emph{Bayesian
Dirichlet} (BD) score:
\begin{equation}
\label{eq:bd}
  \BD(\G, \D; \boldsymbol{\alpha}) = 
  \prod_{i=1}^N \BD(\XPi; \alpha_i) = 
  \prod_{i=1}^N \prod_{j = 1}^{q_i}
    \left[
      \frac{\Gamma(\alpha_{ij})}{\Gamma(\alpha_{ij} + n_{ij})}
      \prod_{k=1}^{r_i} \frac{\Gamma(\aijk + n_{ijk})}{\Gamma(\aijk)}
    \right]
\end{equation}
where $r_i$ is the number of states of $X_i$; $q_i$ is the number of 
configurations of $\PXi$; $n_{ij} = \sum_k n_{ijk}$; 
$\alpha_{ij} = \sum_k \aijk$; and $\boldsymbol{\alpha}$ is the set of the
$\alpha_i$. For $\aijk = 1, \alpha_i = r_i q_i$ we obtain the
K2 score from \citet{k2}; for $\aijk = \fh, \alpha_i = r_i q_i / 2$ we
obtain the BD score with Jeffrey's prior \citep{suzuki16}; and for 
$\aijk = \alpha / (r_i q_i), \alpha_i = \alpha$ we obtain the \emph{Bayesian
Dirichlet equivalent uniform} (BDeu) score from \citet{heckerman}, which is the
most common choice used in score-based algorithms to estimate 
$\Prob(\D \given \G)$.  The corresponding posterior probability estimates of
the $\piijk$ are
\begin{align}
  &\pijk^{(\asi)} 
        = \frac{\aijk + n_{ijk}}{\sum_{k = 1}^{r_i} \aijk + n_{ijk}}
        = \frac{\asi + n_{ijk}}{r_i \asi + n_{ij}}&
  &\text{where $\asi = \frac{\alpha}{r_i q_i}$.}&
\label{eq:postprob}
\end{align}
It can be shown that BDeu is score equivalent 
\citep{chickering}, that is, it takes the same value for DAGs that encode the 
same probability distribution; and that it is the only BD score with this
property \citep[][Theorem 18.4]{koller}. The uniform prior over the parameters
associated with each $\XPi$ has been justified by the lack of prior knowledge and
widely assumed to be non-informative. It is typically used with small imaginary
sample sizes such as $\alpha = 1$ as suggested by \citet{koller} and
\citet{ueno} so that it can be easily dominated by the data.

However, there is an increasing amount of evidence that these assumptions lead
to a prior that is far from non-informative and that has a strong impact on the
quality of the learned DAGs. \citet{silander} showed via simulation that the MAP
DAGs selected using BDeu are highly sensitive to the choice of $\alpha$. Even
for ``reasonable'' values such as $\alpha \in [1, 20]$, they obtained DAGs with
markedly different number of arcs, and they showed that large values of $\alpha$
tend to produce DAGs with more arcs. This is counter-intuitive because larger 
$\alpha$ would normally be expected to result in stronger regularisation and
sparser BNs. \citet{jaakkola} similarly showed that the number of arcs in the 
MAP DAG is determined by a complex interaction between $\alpha$ and $\D$; in the
limits $\alpha \to 0$ and $\alpha \to \infty$ it is possible to obtain both very
sparse and very dense DAGs. (We informally define $\G$ to be \emph{sparse} if 
$|A| = O(N)$, typically with $|A| < 5N$; a \emph{dense} $\G$, on the other hand,
has a relatively large $|A|$ compared to $N$.) In particular, for small values
of $\alpha$ and/or sparse data (that is, discrete data for which we observe a
small subset of the possible combinations of the values of the $X_i$),
$\asi \to 0$ and
\begin{equation}
  \BDeu(\G, \D; \alpha) \to \alpha^{\dEP{\G}}
\label{eq:convergence}
\end{equation}
where $\dEP{\G}$ is the effective number of parameters of the model, defined as
\begin{align*}
  &\dEP{\G} =  \sum_{i = i}^N \dEP{X_i, \G} = \sum_{i = i}^N \left[ \sum_{j = 1}^{q_i} \tilde{r}_{ij} + \tq \right],&
  &\tilde{r}_{ij} = \sum_{k = 1}^{r_i} \bbbone_{\{x > 0\}}(n_{ijk}),
    \tq = \sum_{j = 1}^{q_i} \bbbone_{\{x > 0\}}(n_{ij}),
\end{align*}
where $\tilde{r}_{ij}$ is the number of positive $n_{ijk}$ in the $j$th
configuration of $\PXi$ and $\tq$ is the number of configurations in which at
least one $n_{ijk}$ is positive.

This was then used to prove that
\begin{equation}
  \frac{\Prob(\D \given \Gp)}{\Prob(\D \given \Gm)} = 
  \frac{\BDeu(\XPi \cup X_l; \alpha)}{\BDeu(\XPi; \alpha)}
  \to
  \left\{
  \begin{aligned}
  0 & & \text{if $d_\mathrm{EDF} > 0$}\\
  +\infty & & \text{if $d_\mathrm{EDF} < 0$} 
  \end{aligned}
  \right.
\label{eq:bdeu-ratio}
\end{equation}
for two DAGs $\Gm$ and $\Gp = \Gm \cup \{ X_l \to X_i \}$ that differ only by
the inclusion of a single arc $X_l \to X_i$. The effective degrees of freedom
$d_\mathrm{EDF}$ are defined as $\dEP{\Gp} - \dEP{\Gm}$. The practical
implication of this result is that, if we use the Bayes factor in 
\mref{eq:bdeu-ratio} for structure learning, a large number of zero-cell-counts
will force $d_\mathrm{EDF}$ to be negative, which means the inclusion of
additional arcs is favoured. But that in turn makes $d_\mathrm{EDF}$ even more
negative, quickly leading to overfitting $\G$.

Furthermore, \citet{jaakkola} argued that BDeu can be rather unstable for 
``medium-sized'' data and small $\alpha$, which is a very common scenario. 
\citet{alphastar} approached the problem from a different perspective and
derived an analytic approximation for the ``optimal'' value of $\alpha$ that
maximises predictive accuracy, further suggesting that the interplay between
$\alpha$ and $\D$ is controlled by the skewness of the $\Prob(\XPi)$ and by the
strength of the dependence relationships between the nodes. Skewed
$\Prob(\XPi)$ result in some $\piijk$ being smaller than others, which in turn
makes sparse data sets more likely; hence the problematic behaviour described
in \citet{jaakkola} and reported above. Most of these results have been
analytically confirmed more recently by \citet{ueno,ueno2}.

Finally, \citet{suzuki16} studied the asymptotic properties of BDeu using BD
with Jeffrey's prior as a term of comparison. He found that BDeu is not regular
in the sense that it may learn DAGs in a way that does not respect the maximum
entropy principle \citep{jaynes1,jaynes2} depending on the values of the 
underlying $\piijk$, even if the positivity assumption holds and if $n$ is 
large. This agrees with the observations in \citet{ueno}, who also observed that
BDeu is not necessarily consistent for any finite $n$, but only asymptotically
for $n \to \infty$.

\subsection{The Uniform Graph Prior (U)}

As far as $\Prob(\G)$ is concerned, the most common choice is the uniform (U)
distribution $\Prob(\G) \propto 1$; the space of the DAGs grows
super-exponentially in $N$ \citep{harary} and that makes it extremely difficult
to specify informative priors. Two notable examples are presented in
\citet{csprior} and \citet{mukherjee}. \citet{csprior} described a 
\textit{completed prior} in which they elicitated prior probabilities for a
subset of arcs and completed the prior to cover the remaining arcs with a
discrete uniform distribution. So, if we denote
\begin{align}
\label{eq:banot}
 &\pra = \Prob(\{X_i \rightarrow X_j\} \in A),&
 &\pla = \Prob(\{X_i \leftarrow X_j\} \in A),&
 &\pri = \Prob(\{X_i \rightarrow X_j, \linebreak X_i \leftarrow X_j\} \not\in A)
\end{align}
with $\pra + \pla + \pri = 1$, \citet{csprior} proposed to elicitate the
triplets $(\pra, \pla, \pri)$ for specific pairs of nodes $(X_i, X_j)$, and to
assume $\pra = \pla = \pri = \sfrac{1}{3}$ for the rest. Priors for distinct
$(X_i, X_j)$ were assumed to be independent and thus they can be combined in
\begin{equation*}
  \Prob(\G) \propto \prod_{(i, j)} 
    \pra \bbbone_{\{\{X_i \rightarrow X_j\} \in A\}}(i, j) +
    \pla \bbbone_{\{\{X_i \leftarrow X_j\} \in A\}}(i, j) +
    \pri \bbbone_{\{\{X_i \rightarrow X_j, \linebreak X_i \leftarrow X_j\} \not\in A\}}(i, j).
\end{equation*}
As an alternative, \citet{mukherjee} proposed an informative prior using a
log-linear combination of arbitrary features $f_i(\cdot)$ of $\G$,
\begin{equation*}
  \Prob(\G) \propto \exp\left( \lambda \sum_i w_i f_i(\G) \right),
\end{equation*}
whose relative importance is controlled with some positive weights $w_i$. The
hyperparameter $\lambda$ was used to control the overall strength of the prior,
much like the imaginary sample size in BDeu. While both approaches have been
shown to improve the accuracy of structure learning, they require us to elicit
substantial amounts of information from domain experts, which is notoriously
challenging \citep{madigan}.

The uniform prior is at the opposite end of the spectrum, in that it does not
have any free parameter and therefore does not require any prior elicitation.
In our previous work \citep{ba12}, we explored the first- and second-order
properties of U and we showed that for each possible pair of nodes $(X_i, X_j)$
\begin{align}
\label{eq:ba}
  &\pra = \pla \approx \frac{1}{4} + \frac{1}{4(N-1)}&
  &\text{and}&
  &\pri \approx \frac{1}{2} - \frac{1}{2(N-1)}.
\end{align}
This prior distribution is asymptotically (marginally) uniform over both arc
presence and direction: each arc is present in $\G$ with probability $\fh$ and,
when present, it appears in each direction with probability $\fh$ as 
$N \to \infty$. We also showed that two arcs are correlated if they are incident
on a common node, with
\begin{equation*}
  \COR(A_{ij}, A_{jk}) \approx 
    2\left[\frac{3}{4} - \frac{1}{4(N -1)}\right]^2
     \left[\frac{1}{4} + \frac{1}{4(N - 1)}\right]
\end{equation*}
and $\COR(A_{ij}, A_{kl}) = 0$ otherwise, through exhaustive enumeration of all
possible DAGs for $N \leqslant 7$ and through simulation for larger $N$. This 
suggests that false positives (arcs that are incorrectly included in $\G$) and
false negatives (arcs that are incorrectly excluded from $\G$) can potentially
propagate through $\Prob(\G)$: for instance, if an arc is incorrectly included
in $\G$, then arcs incident on the same head node are now more likely to be
included, which increases the possibility of including further arcs in that part
of $\G$. This in turn may cause further problems in $\Prob(\D \given \G)$, since
the number of parameters of the BN increases combinatorially with the number of
parents of its nodes.

In order to prevent this from happening, many papers in the literature choose to
put a hard limit on the maximum number of parents of each node
\citep[see, for instance,][]{friedman2,sc}. The prior then becomes
\begin{equation*}
  \Prob(\G) \propto \left\{
  \begin{aligned}
  &1& &\text{if $|\PXi| < m$ for all $X_i$} \\
  &0& &\text{otherwise}
  \end{aligned}
  \right.
\end{equation*}
which strongly limits the space of the candidate DAGs by imposing a strong
uniformity constraint on their structure if $m$ is small (that is, all nodes
have about the same small number of parents). Indeed, choosing too small a value
for $m$ has a strong negative impact on the accuracy of structure learning, as
discussed by \citet{elidan}. A ``softer'' alternative would be to use the
classic \textit{variable selection prior} \citep{berger} independently for each
local distribution, that is
\begin{equation*}
  \Prob(\XPi) = \prod_{j \neq i} \pra \bbbone_{\{X_j \in \PXi\}}(j)
\end{equation*}
typically simplified to $\pra = p_i$ or even $\pra = p$ in practical use. This
choice, however, is more problematic to use than the completed prior, because
$\pra$ and $\pla$ are specified independently of each other (they are associated
with different $\XPi$); and because inclusion events are assumed to be
independent for different nodes and for different parents of each node, but they
are not ($X_j \in \PXi$ implies $X_i \not\in \Pi_{X_j}$).

\section{MU+BDs: A Posterior Score Arising from Piecewise Uniform Priors}
\label{sec:bds}

It is clear from the literature review in Section \ref{sec:bdeu} that assuming
uniform priors for $\TT$ and $\G$ can have a negative impact on the quality of
the DAGs learned using U+BDeu. Therefore, we propose a new score with an 
alternative set of assumptions: the \emph{Bayesian Dirichlet sparse} (BDs) 
marginal likelihood with a \emph{marginal uniform} (MU) prior. 

\subsection{The Bayesian Dirichlet Sparse (BDs) Marginal Likelihood}

Firstly, we consider the marginal likelihood BDeu. Starting from \mref{eq:bd}
and \mref{eq:postprob}, we can write it as
\begin{equation}
\label{eq:bdeu}
  \BDeu(\G, \D; \alpha) = \prod_{i=1}^N \BDeu(\XPi; \alpha) = 
  \prod_{i=1}^N \prod_{j = 1}^{q_i}
    \left[
      \frac{\Gamma(r_i \asi)}{\Gamma(r_i \asi + n_{ij})}
      \prod_{k=1}^{r_i} \frac{\Gamma(\asi + n_{ijk})}{\Gamma(\asi)}
    \right].
\end{equation}
If the positivity assumption is violated, the sample size $n$ is small or the
data are sparse, there may be configurations of some $\PXi$ that are not
observed in $\D$. In such cases, $n_{ij} = 0$ and
\begin{equation*}
  \BDeu(\XPi; \alpha) =
    \prod_{j : n_{ij} = 0}
    \left[
    \cancel{
      \frac{\Gamma(r_i \asi)}{\Gamma(r_i \asi)}
      \prod_{k=1}^{r_i} \frac{\Gamma(\asi)}{\Gamma(\asi)}
    }
    \right]
    \prod_{j : n_{ij} > 0}
    \left[
      \frac{\Gamma(r_ i \asi)}{\Gamma(r_i \asi + n_{ij})}
      \prod_{k=1}^{r_i} \frac{\Gamma(\asi + n_{ijk})}{\Gamma(\asi)}
    \right].
\end{equation*}
This implies that the effective imaginary sample size decreases as the number
of unobserved parents configurations increases, since $\sum_{j : n_{ij} > 0}
\sum_k \asi \leqslant \sum_{j, k} \asi = \alpha$. As a result, the posterior
estimates of $\piijk$ gradually converge to the corresponding maximum likelihood
estimates thus favouring overfitting and the inclusion of spurious arcs in $\G$.
(The DAG that maximises the likelihood for any given data set is that that 
corresponds to the saturated model.)

Furthermore, the comparison between DAGs with different numbers of arcs may be
inconsistent because of the different effective imaginary sample sizes used in
the  respective priors. This phenomenon is best illustrated by comparing the
empirical estimator for Shannon entropy \citep{itheory} of $\XPi$, which uses
$\pijk = n_{ijk} / n_{ij}$, with that using the posterior estimates in 
\mref{eq:postprob}. Consider, for instance, two DAGs $\Gm$ and 
$\Gp = \Gm \cup X_l$ as in \mref{eq:bdeu-ratio}. According to the maximum 
entropy principle, we should  prefer $\Gp$ to $\Gm$ if
\begin{align*}
  &\H(\XPi \cup X_l) > \H(\XPi),&
  &\H(\XPi) = - \sum_{j = 1}^{q_i} \sum_{k = 1}^{r_i} \pijk \log  \pijk
\end{align*}
since in that case $\Gp$ captures more information that $\Gm$. For sparse data,
in practice we have
\begin{equation*}
  \H(\XPi) = - \sum_{\jpo} \sum_{k = 1}^{r_i} \pijk \log  \pijk
\end{equation*}
since the terms corresponding to unobserved parent configurations are assumed
to be $0 \log 0 = 0$ for continuity. If we replace the maximum likelihood
estimates $\pijk$ with the corresponding posterior estimates, which are what
BDeu uses to evaluate candidate DAGs, we obtain the following.

\begin{theorem}
In a Bayesian setting, the conditional entropy $\H(\cdot)$ of $\XPi$ given a 
uniform Dirichlet prior with imaginary sample size $\alpha$ over the cell 
probabilities is
\begin{align*}
  &\H(\XPi; \alpha) = - \sum_{\jpo} \sum_{k = 1}^{r_i} \pijk^{(\asi)} \log  \pijk^{(\asi)}&
  &\text{with}&
  &\pijk^{(\asi)} = \frac{\asi + n_{ijk}}{r_i \asi + n_{ij}}.
\end{align*}
and $\H(\XPi; \alpha) < \H(\XPi; \beta)$ if $\alpha < \beta$ and $\XPi$ is not
a uniform distribution.
\label{thm:h-monotonic}
\end{theorem}

Therefore, in the posterior we prefer $\Gp$ to $\Gm$ if 
\begin{equation*}
  \H(\XPi \cup X_l; \alpha) > \H(\XPi; \alpha).
\end{equation*}

Consider, for instance, a non-sparse model $\Gm$ (all $q'_i$ configurations of
$\PXi$ are observed) and a sparse model $\Gp$ ($\tq < q_i$ configurations of
$\PXi \cup X_l$ are observed). For continuity, if $\alpha$ is small and 
$\asi \to 0$ we would expect that
\begin{align*}
  &\H(\XPi \cup X_l; \alpha) \approx \H(\XPi \cup X_l)&
  &\text{and}&
  &\H(\XPi; \alpha) \approx \H(\XPi),
\end{align*}
and that we would be consistent in our choice between $\Gp$ and $\Gm$. Instead
we prefer $\Gp$ over $\Gm$ if
\begin{equation*}
  \H(\XPi \cup X_l; \alpha \tq / q_i) > \H(\XPi; \alpha),
\end{equation*}
where $\H(\XPi \cup X_l; \alpha \tq /q_i) < \H(\XPi \cup X_l; \alpha)$
since $\tq < q_i$. This argument complements that in \citet{suzuki16} in the
context of finite sample sizes and elucidates the underlying reason for the
problematic behaviour of BDeu shown in that paper, which is made clear in the
following theorem. 

\begin{theorem}
Let $\bsi = \beta / (r_i q_i) \to 0$ and let $0 < \alpha < \beta$. Then
\begin{align*}
  &\BDeu(\XPi; \alpha) < \BDeu(\XPi; \beta)&
  &\text{if $\dEP{X_i, \G} > 0$,} \\
  &\BDeu(\XPi; \alpha) = \left(\frac{1}{r_i}\right)^{\tq}&
  &\text{if $\dEP{X_i, \G} = 0$.}
\end{align*}
\label{thm:bdeu-monotonic}
\end{theorem}

Therefore, if $\dEP{X_i, \Gp} > 0$ and $\dEP{X_i, \Gm} > 0$ model selection is
inconsistent because
\begin{equation*}
  \alpha \frac{\tq}{q_i} < \alpha \Rightarrow \left\{
    \begin{aligned}
      &\H(\XPi \cup X_l; \alpha \tq / q_i) < \H(\XPi \cup X_l; \alpha) \\
      &\BDeu(\XPi \cup X_l; \alpha \tq / q_i) < \BDeu(\XPi \cup X_l; \alpha)
    \end{aligned}
  \right.
\end{equation*}
for most finite sample sizes, since we may have a non-zero probability of
comparing a pair of DAGs $\Gp$ and $\Gm$ for which at least one $X_i$ has
$\tq < q_i$. High-dimensional and sparse data sets are especially likely to
produce such inconsistencies because we will evaluate more candidate DAGs in
the course of structure learning and because zero-cell-counts will be more
common. 

We illustrate this phenomenon in the example below for non-singular local 
distributions ($\dEP{X, \Gp} > 0$, $\dEP{X, \Gm} > 0$) below.

\begin{figure}[t]
\begin{center}
  \includegraphics[width=0.9\textwidth]{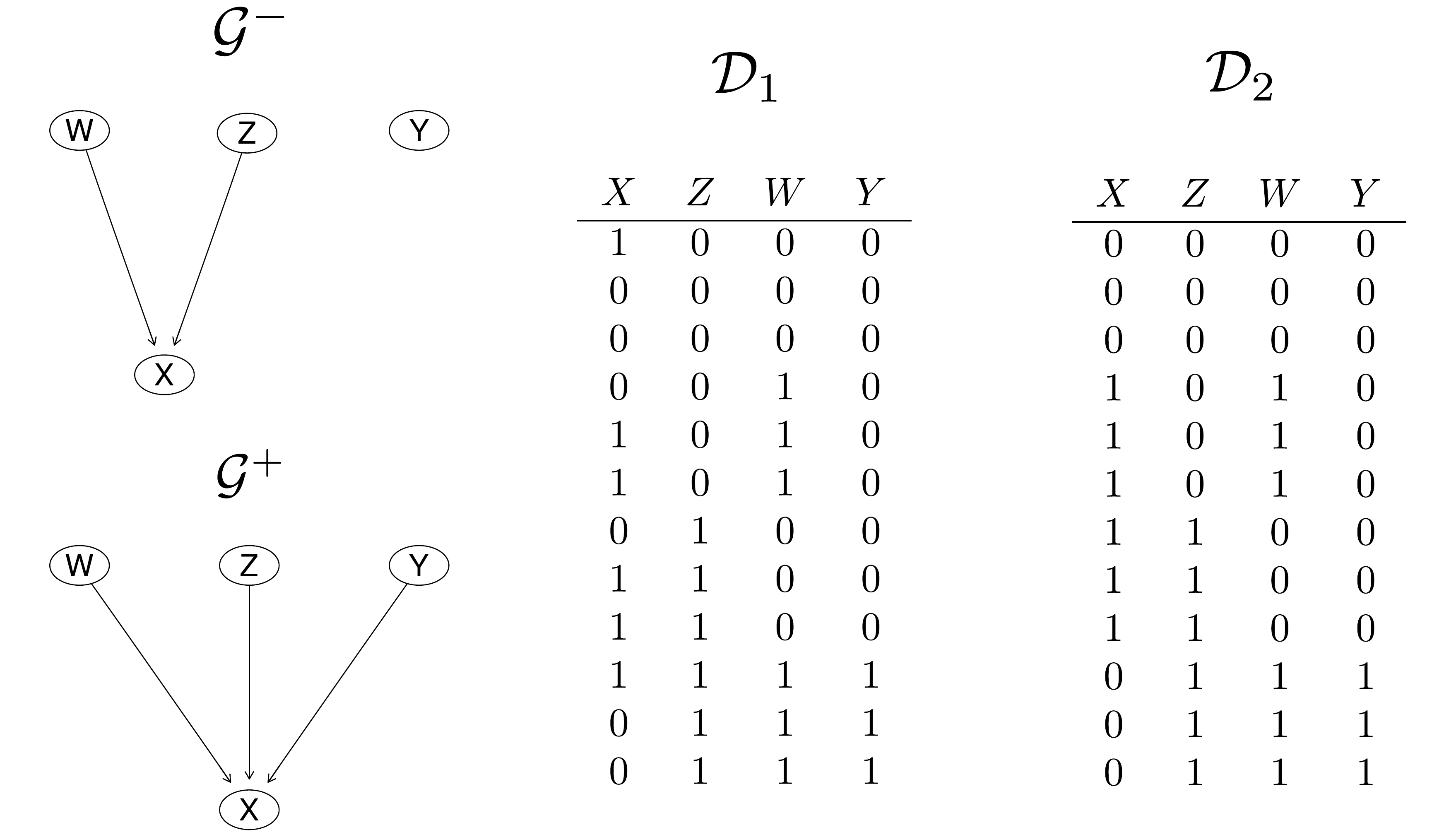}
  \caption{DAGs and data sets used in Examples \ref{ex:nonsingular} and
    \ref{ex:singular}. The DAGs $\Gp$ and $\Gm$ are used in both examples. The
    data set $\D_1$ refers to Example \ref{ex:nonsingular}, while $\D_2$ refers
    to Example \ref{ex:nonsingular}; the former is a modified version of the
    latter, which is originally from \citet{suzuki16}.}
  \label{fig:examples}
\end{center}
\end{figure}

\begin{example}
\label{ex:nonsingular}
Consider a simple example, inspired by that in \citet{suzuki16}, based on the
data set $\D_1$ and the DAGs $\Gm$, $\Gp$ shown in Figure \ref{fig:examples}.
The conditional distributions for $\XP$ in $\Gm$ and for $\XP \cup Y$ in $\Gp$
both have positive empirical entropy. The sample frequencies ($n_{ijk}$) for
$\XP$ are:
\begin{center}
\begin{tabular}{rr|cccc}
    & & \multicolumn{4}{|c}{$Z, W$} \\
    & & $0, 0$ & $1, 0$ & $0, 1$ & $1, 1$ \\
  \hline
  \multirow{2}{*}{$X$}
    & $0$ & $2$ & $1$ & $1$ & $2$ \\
    & $1$ & $1$ & $2$ & $2$ & $1$
\end{tabular}
\end{center}
and those for $\XP \cup Y$ are as follows.
\begin{center}
\begin{tabular}{rr|cccccccc}
    & & \multicolumn{8}{c}{$Z, W, Y$} \\
    & & $0, 0, 0$ & $1, 0, 0$ & $0, 1, 0$ & $1, 1, 0$ 
      & $0, 0, 1$ & $1, 0, 1$ & $0, 1, 1$ & $1, 1, 1$ \\
  \hline
  \multirow{2}{*}{$X$}
    & $0$ & $2$ & $1$ & $1$ & $0$ & $0$ & $0$ & $0$ & $2$ \\
    & $1$ & $1$ & $2$ & $2$ & $0$ & $0$ & $0$ & $0$ & $1$
\end{tabular}
\end{center}
Therefore, the marginal likelihood for $\XP$ is estimated from a contingency
table in which all parents configurations are observed in the data. On the other
hand, we only observe $4$ out of $8$ parents configurations in the contingency 
table for $\XP \cup Y$.

Even though $\XP$ and $\XP \cup Y$ have the same empirical entropy,
\begin{equation*}
  \H(\XP) = \H(\XP \cup Y) = 4 \left[
    - \frac{1}{3} \log \frac{1}{3} - 
      \frac{2}{3} \log \frac{2}{3}\right] = 2.546;
\end{equation*}
if $\alpha = 1$, $\asi = \sfrac{1}{8}$ for $\Gm$ and $\asi = \sfrac{1}{16}$ for
$\Gp$ , so the posterior entropies are different:
\begin{align*}
  \H(\XP; 1) &= 4 \left[
    - \frac{1 + \sfrac{1}{8}}{3 + \sfrac{1}{4}} \log \frac{1 + \sfrac{1}{8}}{3 + \sfrac{1}{4}} - 
      \frac{2 + \sfrac{1}{8}}{3 + \sfrac{1}{4}} \log \frac{2 + \sfrac{1}{8}}{3 + \sfrac{1}{4}} \right] = 2.580,\\
  \H(\XP \cup Y; 1) &= 4 \left[
    - \frac{1 + \sfrac{1}{16}}{3 + \sfrac{1}{8}} \log \frac{1 + \sfrac{1}{16}}{3 + \sfrac{1}{8}} - 
      \frac{2 + \sfrac{1}{16}}{3 + \sfrac{1}{8}} \log \frac{2 + \sfrac{1}{16}}{3 + \sfrac{1}{8}} \right] = 2.564. 
\end{align*}
Therefore, $\Gm$ would be preferred over $\Gp$, and that is the decision that
is reached using BDeu:
\begin{align*}
  \BDeu(\XP; 1) &= 
    \left( \frac{\Gamma(\sfrac{1}{4})}{\Gamma(\sfrac{1}{4}+3)} 
    \left[ \frac{\Gamma(\sfrac{1}{8} + 2)}{\Gamma(\sfrac{1}{8})} \cdot 
           \frac{\Gamma(\sfrac{1}{8} + 1)}{\Gamma(\sfrac{1}{8})} \right] \right)^4
              = 3.906\e{-7}, \\
  \BDeu(\XP \cup Y; 1) &= 
    \left( \frac{\Gamma(\sfrac{1}{8})}{\Gamma(\sfrac{1}{8}+3)} 
    \left[ \frac{\Gamma(\sfrac{1}{16} + 2)}{\Gamma(\sfrac{1}{16})} \cdot 
           \frac{\Gamma(\sfrac{1}{16} + 1)}{\Gamma(\sfrac{1}{16})} \right] \right)^4
                     = 3.721\e{-8}.
\end{align*}
Like the posterior entropy, we note that BDeu takes different values for two
local distributions, $\XP$ and $\XP \cup Y$, that encode exactly the same 
information.
\end{example}

Structure learning is inconsistent also if at least one of the DAGs implies a
singular and sparse local distribution (say, $\dEP{X, \Gp} = 0$ below) since in
that case BDeu converges to a constant that does not depend on the data as
$\asi \to 0$, unlike the posterior conditional entropy.

\begin{example}
\label{ex:singular}
Consider the second data set $\D_2$ in Figure \ref{fig:examples}, originally
from \citet{suzuki16}, and the same DAGs $\Gm$ and $\Gp$. The sample frequencies
for $\XP$ are:
\begin{center}
\begin{tabular}{rr|cccc}
    & & \multicolumn{4}{c}{$Z, W$} \\
    & & $0, 0$ & $1, 0$ & $0, 1$ & $1, 1$ \\
  \hline
  \multirow{2}{*}{$X$}
    & $0$ & $3$ & $0$ & $0$ & $3$ \\
    & $1$ & $0$ & $3$ & $3$ & $0$
\end{tabular}
\end{center}
and those for $\XP \cup Y$ are as follows.
\begin{center}
\begin{tabular}{rr|cccccccc}
    & & \multicolumn{8}{c}{$Z, W, Y$} \\
    & & $0, 0, 0$ & $1, 0, 0$ & $0, 1, 0$ & $1, 1, 0$
      & $0, 0, 1$ & $1, 0, 1$ & $0, 1, 1$ & $1, 1, 1$ \\
  \hline
  \multirow{2}{*}{$X$}
    & $0$ & $3$ & $0$ & $0$ & $0$ & $0$ & $0$ & $0$ & $3$ \\
    & $1$ & $0$ & $3$ & $3$ & $0$ & $0$ & $0$ & $0$ & $0$
\end{tabular}
\end{center}
The empirical entropy of $X$ given its parents is equal to zero for both $\Gp$
and $\Gm$, since the value of $X$ is completely determined by the configurations
of its parents in both DAGs. Again, the posterior entropies for $\Gp$ and $\Gm$
differ:
\begin{align*}
  \H(\XP; 1) &= 4 \left[
    - \frac{0 + \sfrac{1}{8}}{3 + \sfrac{1}{4}} \log \frac{0 + \sfrac{1}{8}}{3 + \sfrac{1}{4}} - 
      \frac{3 + \sfrac{1}{8}}{3 + \sfrac{1}{4}} \log \frac{3 + \sfrac{1}{8}}{3 + \sfrac{1}{4}} \right] = 0.652,\\
  \H(\XP \cup Y; 1) &= 4 \left[
    - \frac{0 + \sfrac{1}{16}}{3 + \sfrac{1}{8}} \log \frac{0 + \sfrac{1}{16}}{3 + \sfrac{1}{8}} - 
      \frac{3 + \sfrac{1}{16}}{3 + \sfrac{1}{8}} \log \frac{3 + \sfrac{1}{16}}{3 + \sfrac{1}{8}} \right] = 0.392. 
\end{align*}
However, BDeu with $\alpha = 1$ yields
\begin{align*}
  \BDeu(\XP; 1) &= 
    \left( \frac{\Gamma(\sfrac{1}{4})}{\Gamma(\sfrac{1}{4}+3)} 
    \left[ \frac{\Gamma(\sfrac{1}{8} + 3)}{\Gamma(\sfrac{1}{8})} \cdot 
           \cancel{\frac{\Gamma(\sfrac{1}{8})}{\Gamma(\sfrac{1}{8})}} \right] \right)^4
    = 0.0326, \\
  \BDeu(\XP \cup Y; 1) &= 
    \left( \frac{\Gamma(\sfrac{1}{8})}{\Gamma(\sfrac{1}{8}+3)} 
    \left[ \frac{\Gamma(\sfrac{1}{16} + 3)}{\Gamma(\sfrac{1}{16})} \cdot 
           \cancel{\frac{\Gamma(\sfrac{1}{16})}{\Gamma(\sfrac{1}{16})}} \right] \right)^4 
    = 0.0441,
\end{align*}
preferring $\Gp$ over $\Gm$ even though the additional arc $Y \to X$ does not
provide any additional information on the distribution of $X$, and even though
$4$ out of $8$ conditional distributions in $\XP \cup Y$ are not observed at all
in the data. In fact, both the empirical and the posterior entropies would lead
to selecting $\Gm$ over $\Gp$ in this example.

\end{example}

To address these undesirable features of BDeu we propose to replace $\asi$ in
\mref{eq:bdeu} with
\begin{align}
\label{eq:newprior}
  &\ta_i = \left\{
    \begin{aligned}
      &\alpha / (r_i \tq)& & \text{if $n_{ij} > 0$} \\
      &0                         & & \text{otherwise.}
    \end{aligned}
  \right.& &\text{where}&
  &\tq = \{ \text{number of $\PXi$ such that $n_{ij} > 0$} \}.
\end{align}
Note that \mref{eq:newprior} is still piece-wise uniform, but now
$\sum_{j : n_{ij} > 0} \sum_k \ta_i = \alpha$ so the effective imaginary sample
size is equal to $\alpha$ even for sparse data. Intuitively, we are defining a
uniform prior just on the conditional distributions we can estimate from $\D$,
thus moving from a fully Bayesian to an empirical Bayes score. Plugging 
\mref{eq:newprior} in \mref{eq:bd} we obtain BDs:
\begin{equation}
  \BDs(\XPi; \alpha) =
    \prod_{j : n_{ij} > 0}
    \left[
      \frac{\Gamma(r_i \ta_i)}{\Gamma(r_i \ta_i + n_{ij})}
      \prod_{k=1}^{r_i} \frac{\Gamma(\ta_i + n_{ijk})}{\Gamma(\ta_i)}
    \right],
\label{eq:bds}
\end{equation}
and we can write
\begin{equation}
  \BDs(\XPi; \alpha) = \BDeu(\XPi; \alpha q_i / \tq).
\label{eq:uneven}
\end{equation}

The relationship between BDeu and BDs for small and large imaginary sample 
sizes and for large sample sizes is as follows.

\begin{theorem}
If $\asi \to 0$, then
\begin{equation*}
  \BDs(\XPi; \alpha) =
    \BDeu(\XPi; \alpha) \cdot \left( \frac{q_i}{\tq} \right)^{\dEP{X_i, \G}}.
\end{equation*}
\label{thm:bdlink}
\end{theorem}

\begin{theorem}
BDs is equivalent to BDeu if at least one the following conditions holds:
\begin{enumerate}
  \item $n \to \infty$ and the positivity assumption holds; \label{thm:eq:one}
  \item $\alpha \to \infty$. \label{thm:eq:two}
\end{enumerate}
\label{thm:asymeq}
\end{theorem}

We can interpret $q_i / \tq$ as an adaptive regularisation hyperparameter that
corrects for $\XPi$ that are not fully observed in $\D$, which typically
correspond to $X_i$ with a large number of incoming arcs; or equivalently as a
finite sample correction for BDeu. We can also link BDs to posterior entropy by
defining
\begin{align*}
  &\tilde{\H}(\XPi; \alpha) = - \sum_{\jpo} \sum_{k = 1}^{r_i} \pijk^{(\tai)} \log  \pijk^{(\tai)}&
  &\text{with}&
  &\pijk^{(\tai)} = \frac{\tai + n_{ijk}}{r_i \tai + n_{ij}}.
\end{align*}
as we did for BDeu in Theorem \mref{thm:h-monotonic}. Since now
the imaginary sample size is equal to $\alpha$ even when some $n_{ij} = 0$, model
selection is again consistent as we can see by revisiting Examples 
\ref{ex:nonsingular} and \ref{ex:singular}.

\setcounter{example}{0}

\begin{example}{\bf (Continued)}
Consider again the data set $\D_1$. BDs does not suffer from the bias arising
from $\tq < q_i$ and it correctly assigns the same score to both networks
\begin{equation*}
  \BDs(\XP; 1) = \BDs(\XP \cup Y; 1) = 3.906\e{-7}
\end{equation*}
following the maximum entropy principle, and
\begin{equation*}
  \tilde{\H}(\XP; 1) = \tilde{\H}(\XP \cup Y; 1) = 2.580.
\end{equation*}

\begin{figure}[t!]
\begin{center}
  \includegraphics[width=0.9\textwidth]{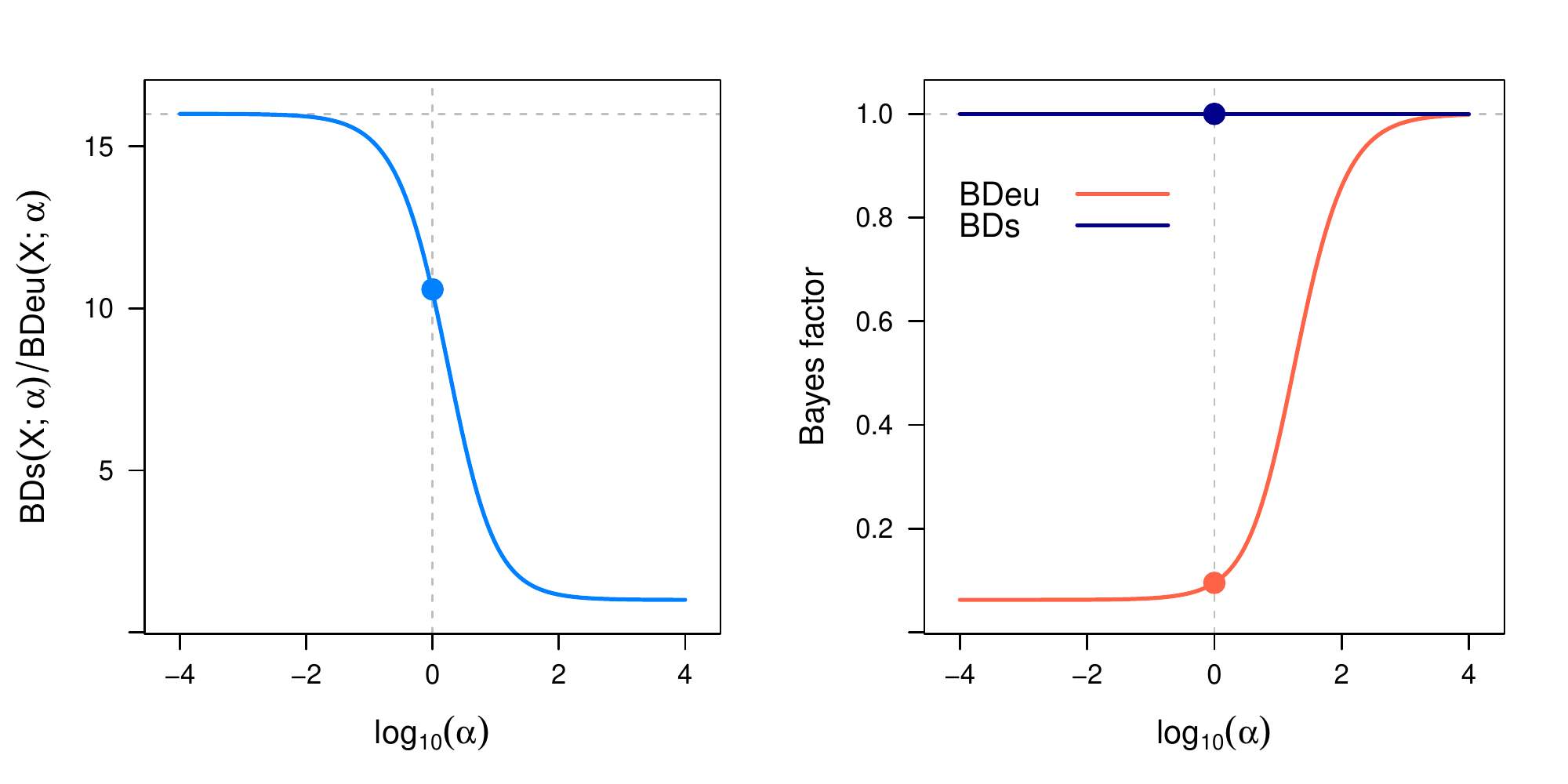}
  \caption{The ratio between $\BDs(\XP \cup Y; \alpha)$ and
    $\BDeu(\XP \cup Y; \alpha)$ (left panel) from Example \ref{ex:nonsingular};
    and the Bayes factors for $\Gp$ versus $\Gm$ computed using BDeu and BDs 
    (right panel; in orange and dark blue, respectively). The bullet points
  correspond to the values observed for $\alpha = 1$.}
  \label{fig:limits2}
\end{center}
\end{figure}

We can also verify that the limit results in Theorems 
\ref{thm:bdlink} and \ref{thm:asymeq} hold for $\Gp$ (BDeu and BDs are
identical for $\Gm$), as shown in the left panel Figure \ref{fig:limits2}:
\begin{align*}
  &\lim_{\asi \to 0} 
   \frac{\BDs(\XP \cup Y; \alpha)}
        {\BDeu(\XP \cup Y; \alpha)} 
     = \left(\frac{8}{4}\right)^{\dEP{X, \Gp}} = 16,&
  &\lim_{\asi \to \infty}
   \frac{\BDs(\XP \cup Y; \alpha)}{\BDeu(\XP \cup Y; \alpha)} = 1,
\end{align*}
where $\dEP{X, \Gp} = 8 - 4 = 4$. Furthermore, 
\begin{align*}
  &\H(\XP) < \H(\XP; \alpha)& &\text{and}&
  &\H(\XP \cup Y) < \H(\XP \cup Y; \alpha) 
\end{align*}
as proved in Theorem \ref{thm:h-monotonic}, and
\begin{align*}
  \BDs(\XP \cup Y; \alpha) = \BDeu(\XP \cup Y; 2 \alpha) 
                           > \BDeu(\XP \cup Y; \alpha) 
\end{align*}
following Theorem \ref{thm:bdeu-monotonic}.
\end{example}

\begin{example}{\bf (Continued)}
Consider again the data set $\D_2$. BDs yields
\begin{equation*}
  \BDs(\XP; 1) = \BDs(\XP \cup Y; 1)
            = 0.0326,
\end{equation*}
which leads to $\Gp$ being discarded in favour of $\Gm$ since its score is
not strictly greater. Similarly,
\begin{equation*}
  \tilde{\H}(\XP; 1) = \tilde{\H}(\XP \cup Y; 1) = 0.652.
\end{equation*}

\begin{figure}[t]
\begin{center}
\includegraphics[width=0.9\textwidth]{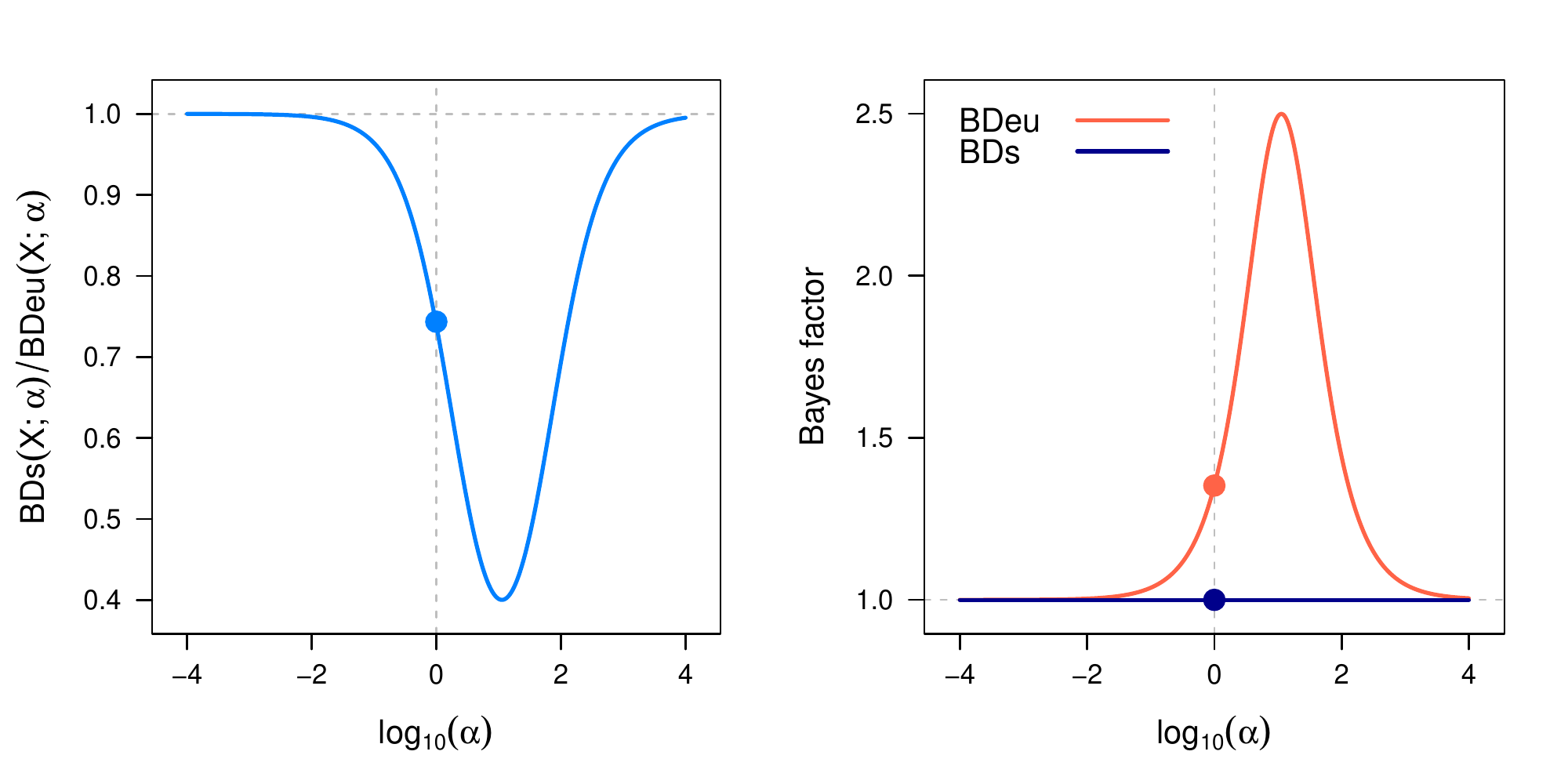}
\caption{The ratio between $\BDs(\XP \cup Y; \alpha)$ and
  $\BDeu(\XP \cup Y; \alpha)$ (left panel) from Example \ref{ex:singular};
  and the Bayes factors for $\Gp$ versus $\Gm$ computed using BDeu and BDs
  (right panel; in orange and dark blue, respectively). The bullet points
  correspond to the values observed for $\alpha = 1$.}
\label{fig:limits1}
\end{center}
\end{figure}

We can also verify that the limit results in Theorems 
\ref{thm:bdlink} and \ref{thm:asymeq} hold for $\Gp$ (BDeu and BDs are
identical for $\Gm$):
\begin{align*}
  &\lim_{\asi \to 0} \frac{\BDs(\XP \cup Y; \alpha)}{\BDeu(\XP \cup Y; \alpha)} = \left(\frac{4}{4}\right)^{\dEP{X, \Gp}} = 1&
  &\lim_{\asi \to \infty} \frac{\BDs(\XP \cup Y; \alpha)}{\BDeu(\XP \cup Y; \alpha)} = 1,
\end{align*}
where $\dEP{X, \Gp} = 4 - 4 = 0$. This is illustrated in the left panel of
Figure \ref{fig:limits1}.

\end{example}

If the positivity assumption holds, we will eventually observe all parents 
configurations in the data and thus $\BDs(\XPi; \alpha) \to \BDeu(\XPi; \alpha)$
as $n \to \infty$. Note, however, that BDs is not score equivalent for finite
$n$ unless all  $n_{ij} > 0$. A numeric example is given below.

\begin{example}
  Consider two binary variables $X$ and $Y$ with data $\D_3$ as follows:
  \begin{center}
  \begin{tabular}{rr|cc}
      & & \multicolumn{2}{c}{$Y$} \\
      & & $0$ & $1$ \\
    \hline
    \multirow{2}{*}{$X$}
      & $0$ & $0$ & $2$ \\
      & $1$ & $0$ & $5$ 
  \end{tabular}
  \end{center}
  If $\alpha = 1$, $\G_1 = \{ Y \to X \}$ and  $\G_2 = \{ X \to Y \}$, then
  \begin{multline*}
  \BDs(\G_1, \D_3; 1) = \\ =
  \left[\frac{\Gamma(1)}{\Gamma(1 + 7)}
        \cancel{\frac{\Gamma(\fh + 0)}{\Gamma(\fh)}}
        \frac{\Gamma(\fh + 7)}{\Gamma(\fh)}\right]
  \left[\frac{\Gamma(1)}{\Gamma(1 + 7)}
        \frac{\Gamma(\fh + 2)}{\Gamma(\fh)}
        \frac{\Gamma(\fh + 5)}{\Gamma(\fh)}\right]
  = 0.0009,
  \end{multline*}
  \begin{multline*}
  \BDs(\G_2, \D_3; 1) =
    \left[\frac{\Gamma(1)}{\Gamma(1 + 7)}
          \frac{\Gamma(\fh + 2)}{\Gamma(\fh)}
          \frac{\Gamma(\fh + 5)}{\Gamma(\fh)}\right] \cdot \\ \cdot
    \left[\frac{\Gamma(\fh)}{\Gamma(\fh + 2)}
          \frac{\Gamma(\fh)}{\Gamma(\fh + 5)}
          \cancel{\frac{\Gamma(\sfrac{1}{4} + 0)}{\Gamma(\sfrac{1}{4})}}
          \cancel{\frac{\Gamma(\sfrac{1}{4} + 0)}{\Gamma(\sfrac{1}{4})}}
          \frac{\Gamma(\sfrac{1}{4} + 2)}{\Gamma(\sfrac{1}{4})}
          \frac{\Gamma(\sfrac{1}{4} + 5)}{\Gamma(\sfrac{1}{4})}\right]
    = 0.0006;
  \end{multline*}
  as a term of comparison the empty DAG $\G_0$ has $\BDs(\G_0, \D_3) = 0.0009$.
\end{example}

In the general case, BDs breaks the score equivalence condition in
\citet{heckerman} because it potentially associates a different imaginary sample
size to each node as shown in \mref{eq:uneven}. 

Since \citet{jaakkola} showed that BDeu favours the inclusion of spurious arcs
for sparse $\XPi$, BDs should lead to sparser DAGs and reduce overfitting; we 
have seen some evidence of that in Example \ref{ex:singular}. The difference in
their model selection choices can be characterised by the Bayes factor computed
using BDs with that computed using BDeu. If we apply Theorem \ref{thm:bdlink} to
\mref{eq:bdeu-ratio} we obtain:
\begin{equation}
  \underbrace{\frac{\BDs(\XPi \cup X_l; \alpha)}
                   {\BDs(\XPi; \alpha)}}_{\text{Bayes factor from BDs}} =
  \underbrace{\frac{(q_i / \tq)^{\dEP{X_i, \Gp}}}
                   {(q'_i / \tilde{q}'_i)^{\dEP{X_i, \Gm}}}}_{\text{implicit prior}} \cdot 
  \underbrace{\frac{\BDeu(\XPi \cup X_l; \alpha)}
                   {\BDeu(\XPi; \alpha)}}_{\text{Bayes factor from BDeu}}.
\label{eq:relationship}
\end{equation}
The first term on the right-hand side can be interpreted as a ratio of empirical
prior probabilities implied by BDs, which are a function of the sparsity of $\Gp$
and $\Gm$ in the parameter space. If all parents configurations are observed in
both models ($q_i = \tq$ and $q'_i = \tilde{q}'_i$), then the prior ratio term
vanishes for small $\alpha$. If at least one of $\Gp$ and $\Gm$ contains
unobserved configurations of $\PXi$, then BDs leads to different choices than
BDeu as shown below.

\begin{theorem}
Let $\Gp$ and $\Gm$ be two DAGs differing from a single arc $X_l \to X_i$, and
let $\asi \to 0$. Then the Bayes factor computed using BDs corresponds to the
Bayes factor computed using BDeu weighted by the following implicit prior ratio:
\begin{equation*}
  \frac{\Prob(\Gp)}{\Prob(\Gm)} = 
  \frac{(q_i / \tq)^{\dEP{X_i, \Gp}}}{(q'_i / \tilde{q}'_i)^{\dEP{X_i, \Gm}}}.
\end{equation*}
and from \mref{eq:convergence} and \mref{eq:bdeu-ratio} can be written as
\begin{align*}
  \frac{\BDs(\XPi \cup X_l; \alpha)}{\BDs(\XPi; \alpha)} &=
  \frac{(q_i / \tq)^{\dEP{X_i, \Gp}} \alpha^{\dEP{\Gp}}}
       {(q'_i / \tilde{q}'_i)^{\dEP{X_i, \Gm}} \alpha^{\dEP{\Gm}}} \\
  &\to
  \left\{
  \begin{aligned}
  0 & & \text{if $d_\mathrm{EDF} > -\log_\alpha (\Prob(\Gp) / \Prob(\Gm))$}\\
  +\infty & & \text{if $d_\mathrm{EDF} < -\log_\alpha (\Prob(\Gp) / \Prob(\Gm))$} 
  \end{aligned}
  \right..
\end{align*}
\label{thm:threshold}
\end{theorem}

As we can see in the right panels of Figures \ref{fig:limits2} and 
\ref{fig:limits1}, the Bayes factor constructed from BDeu can assume very 
different values for the $\Gp$ and $\Gm$ in Examples \ref{ex:nonsingular} and
\ref{ex:singular} depending on the value of $\alpha$, which is not the case for
the Bayes factor constructed from BDs. Considering that in a Bayesian setting
we rely on the ordering Bayes factors to choose the best candidate DAG in each
step of structure learning, and that Bayes factors constructed from BDeu will
vary in different ways for different pairs of DAGs even for the same $\alpha$,
this is highly problematic for BDeu and a strong point in favour of BDs.

\subsection{The Marginal Uniform (MU) Graph Prior}

We now propose a modified prior over for $\G$ with the same aims. We again start
from the consideration that score-based structure learning algorithms typically
generate new candidate DAGs by a single arc addition, deletion or reversal. So,
for example
\begin{equation}
\label{eq:hcstep}
  \Prob(\Gp \given \D) > \Prob(\Gm \given \D)
  \Rightarrow \text{accept $\Gp$ and discard $\Gm$}.
\end{equation}
When using the U prior we can rewrite the left-hand of \mref{eq:hcstep} as
\begin{equation}
\label{eq:hcstep2}
  \frac{\Prob(\Gp \given \D)}{\Prob(\Gm \given \D)} =
  \cancel{\frac{\Prob(\Gp)}{\Prob(\Gm)}}
  \frac{\Prob(\D \given \Gp)}{\Prob(\D \given \Gm)} > 1. 
\end{equation}
The fact that U always simplifies is equivalent to assigning equal probabilities
to all possible states of an arc (subject to the acyclicity constraint), say
$\pra = \pla = \pri = \sfrac{1}{3}$ using the notation in \mref{eq:banot}. The
probabilities in \mref{eq:hcstep2} are different from the marginal probabilities
in \mref{eq:ba} because they represent the conditional inclusion probabilities
for an arc $X_l \rightarrow X_i$ given the rest of $\Gp$, which is kept fixed as
the ``baseline'' $\Gm$ DAG. As a result, U favours the inclusion of new arcs as
$\pra + \pla = \sfrac{2}{3}$. (The same is true for the completed prior from
\citet{csprior}, at least for the pairs of nodes for which we do not elicitate
prior probabilities.) Since in \citet{ba12} we also showed that arcs incident on
a common node are correlated and may favour each other's inclusion, U may then
contribute to overfitting.

Therefore, we introduce the \emph{marginal uniform} (MU) prior, in which we
assume an independent prior for each arc as in \citet{csprior}, with
probabilities
\begin{align*}
  &\pra = \pla = \frac{\beta}{2}&
  &\text{and}& &\pri = 1 - \beta,& &\beta \in (0, 1)
\end{align*}
for all arcs. For $\beta = \fh$ we obtain
\begin{align*}
  &\pra = \pla = \frac{1}{4}&
  &\text{and}& &\pri = \frac{1}{2}& &\text{for all $i \neq j$}
\end{align*}
as in \citet{ba12}, which is essentially the \textit{median-probability prior}
described in \citet{barbieri}. These assumptions make MU computationally trivial
to use: the ratio of the prior probabilities is $\fh$ for arc addition, $2$ for
arc deletion and $1$ for arc reversal, for all arcs. Furthermore, arc inclusion
now has the same prior probability as arc exclusion ($\pra + \pla = \pri = \fh$)
and arcs incident on a common node are no longer correlated, thus limiting 
overfitting and preventing the inclusion of spurious arcs to propagate. However,
the marginal distribution for each arc is the same as in \mref{eq:ba} for large
$N$, hence the name ``marginal uniform''.

While the median-probability prior has been shown to result in better predictive
power than the uniform prior \citep{barbieri,berger}, we note that its expected
number of arcs is
\begin{equation*}
  \E(|A|) = \frac{N(N - 1)}{2}\beta = O(N^2 \beta)
\end{equation*}
which at least in principle would encourage the selection of dense DAGs, leaving
it to the marginal likelihood to penalise overly complex BNs. Therefore, we may
want to consider
\begin{equation}
  \beta = cN \cdot \frac{2}{N(N - 1)} = \frac{2c}{N - 1}
\end{equation}
for some small positive constant $c$, so that $\E(|A|) = O(N)$. That constant 
should be greater or equal to $1$, since we need more arcs than nodes to have
a fully connected DAG; $c = 1$ implies that expected DAG in the prior is a 
tree. In practice, we may want to consider only values smaller than $5$ as in
our experience real-world BNs typically have fewer than $5N$ arcs.

\section{Simulation Study}
\label{sec:sim}

\begin{table}[b]
\begin{center}
  \begin{tabular}{|l|r|r|r|c|l|r|r|r|}
  \cline{1-4} \cline{6-9} 
  network    & $N$   & $|A|$ & $p$        & & network    & $N$   & $|A|$ & $p$      \\
  \cline{1-4} \cline{6-9} 
  ALARM      & $37$  & $46$  & $509$      & & HEPAR 2    & $70$  & $123$ & $1453$   \\
  ANDES      & $223$ & $338$ & $1157$     & & INSURANCE  & $27$  & $52$  & $984$    \\
  CHILD      & $20$  & $25$  & $230$      & & PATHFINDER & $135$ & $200$ & $77155$  \\
  DIABETES   & $413$ & $602$ & $429409$   & & PIGS       & $442$ & $592$ & $5618$   \\
  HAILFINDER & $56$  & $66$  & $2656$     & & WATER      & $32$  & $66$  & $10083$  \\
  \cline{1-4} \cline{6-9} 
  \end{tabular}
  \caption{Reference BNs from the BN repository \citep{bnrepository} with the
     respective numbers of nodes ($N$), numbers of arcs ($|A|$) and numbers of
     parameters ($p = |\Theta|$).}
  \label{tab:networks}
\end{center}
\end{table}

We assessed BDs and MU on a set of $10$ reference BNs (Table \ref{tab:networks})
covering a wide range of $N$ ($20$ to $442$), $p = |\Theta|$ ($230$ to $77$K)
and number of arcs $|A|$ ($25$ to $602$). For each BN:
\begin{enumerate}
  \item We generated $20$ training samples of size $\sfrac{n}{p} = 0.1$, $0.2$,
     $0.5$, $1.0$, $2.0$, and $5.0$ (to allow for meaningful comparisons between
    BNs with such different $N$ and $p$).
  \item We learned $\G$ using U+BDeu, U+BDs, MU+BDeu and MU+BDs with
    $\alpha = 1, 5, 10$ and $\beta = \fh$, $\beta = 2c/(N - 1)$ for $c = 1, 2, 5$
    on each sample. For U + BDeu we also considered the optimal $\alpha$ from
    \citet{alphastar}, denoted $\alpha_S$. In addition, we considered BIC as a
    term of comparison, since $\BIC \to \log\BDeu$ as $n \to \infty$. 
  \item We measured the performance of different scoring strategies in terms of:
    \begin{itemize}
      \item the quality of the learned DAG using the SHD distance \citep{mmhc} 
        from the $\GR$ of the reference BN;
      \item the number of arcs compared to $|\AR|$ in $\GR$;
      \item predictive accuracy, computing the log-likelihood on a test set of
        size $10$k as an approximation of the corresponding Kullback-Leibler
        distance.
    \end{itemize}
    The significance of the difference between different scoring strategies
    for these quantities was assessed using paired t-tests \citep{wasserman}
    with a $p$-value threshold of $0.01$, accompanied by the number of 
    combinations of BNs and $\sfrac{n}{p}$ (out of $60$) in which the scoring
    strategy of interest is better.
\end{enumerate}
For parameter learning, we used Dirichlet posterior estimates and $\alpha = 1$
as suggested in \cite{koller}. All simulations were performed using the
hill-climbing implementation in the \emph{bnlearn} R package \citep{jss09}, which
provides several options for structure learning, parameter learning and inference
on BNs (including the proposed MU and BDs). Since $\alpha = 5$ produced
performance measures that are always in between those for $\alpha = 1$ and
$\alpha = 10$, we omit its discussion for brevity.

\subsection{Comparing U+BDeu with MU+BDs}
\label{sec:sim-bds}

Firstly, we compare U+BDeu, U+BDs, MU+BDeu and MU+BDs for $\beta = \fh$, that
is, taking the median-probability prior for MU. SHD distances are reported in
Table \ref{tab:shd}, from which we observe the following.
\begin{itemize}
  \item MU+BDs outperforms U+BDeu in terms of SHD ($p$-value: $1.8\e{-4}$ for
    $\alpha = 1$, $4.5\e{-9}$ for $\alpha = 10$; $59/60$ and $58/60$ simulations
    respectively), and it is the best score overall in $40/60$ simulations.

  \item BIC also outperforms U+BDeu for $\alpha = 10$ ($p$-value: $2\e{-4}$;
    $54/60$); but it does not significantly outperform U+BDeu for $\alpha = 1$
    ($p$-value: $0.11$; $40/60$). Furthermore, BIC does not outperform MU+BDs
    for either $\alpha = 1$ ($p$-value: $0.97$; $17/60$) or $\alpha = 10$ 
    ($p$-value: $0.11$, $34/60$). BIC is the best score overall in only $13/60$
    simulations.

  \item The improvement in SHD given by using BDs instead of BDeu and by using
    MU instead of U appears to be somewhat non-additive; MU+BDs in most cases
    has the same or nearly the same SHD as the best between U+BDs and MU+BDeu.
    MU+BDs does not significantly outperform either MU+BDeu for for $\alpha = 1$
    ($p$-value: $0.043$; $4/60$) or U+BDs ($p$-value: $0.058$ for $\alpha = 1$,
    $0.029$ for $\alpha = 10$; $17/60$, $4/60$). However, it does 
    significantly outperform MU+BDeu for $\alpha = 10$ ($p$-value: $3.7\e{-6}$;
    $53/60$).

    Overall, MU+BDs outperforms the best between MU+BDeu and U+BDs for both
    $\alpha = 1$ ($p$-value: $0.009$) and $\alpha = 10$ ($p$-value: $8.2\e{-5}$);
    so we recommend it over other combinations of graph priors and marginal
    likelihoods.

    Finally, we note that MU+BDeu is tied with MU+BDs for the best SHD more
    often than U+BDs ($25/27$ vs $2/13$) which suggests improvements in SHD can
    be attributed more to the use of MU than that of BDs.

  \item For U+BDeu, $\alpha = 1$ very often results in a lower SHD than $\alpha_s$
    ($p$-value: $7.4\e{-6}$; $53/60$) and $\alpha = 10$ ($p$-value: $1\e{-5}$;
    $60/60$), which is in agreement with \citet{ueno}. 

  \item For all scoring strategies we observe a strong ($\leqslant -0.85$)
    negative correlation between SHD and $\log(\sfrac{n}{p})$ for $7$ out of
    $10$ BNs, which suggests that the SHD distance from $\GR$ decreases linearly
    as $\log(\sfrac{n}{p})$ increases. This observation complements the results
    on the decay of the probability of single-step structural learning error
    presented in \citet{zuk}. Interestingly, correlation is positive for two BNs
    -- DIABETES and PIGS -- as we can see that SHD is increasing in 
    $\log\sfrac{n}{p}$ in Table \ref{tab:shd}. We have no explanation for this
    phenomenon, which represents an interesting direction for future research.

\end{itemize}
The higher SHD for U+BDeu seems to be a consequence of the higher number of arcs
present in the learned DAGs, shown in Table \ref{tab:narcs}. 
\begin{itemize}
  \item MU+BDs learns significantly fewer arcs than U+BDeu for both $\alpha = 1$
    ($p$-value: $2.4\e{-8}$; $57/60$) and $\alpha = 10$ ($p$-value: $3.5\e{-13}$;
    $60/60$). MU+BDs learns too many arcs (that is, the ratio with 
    $|\AR|$ is greater than $1$) in $23/60$ and $30/60$ simulations,
    as opposed to $32/60$ and $56/60$ for U+BDeu.

  \item The same is true for BIC, which learns fewer arcs than U+BDeu ($p$-value:
    $1.4\e{-10}$ for $\alpha = 1$, $6.8\e{-14}$ for $\alpha = 10$; $59/60$ and
    $60/60$) and learns too many arcs in only $18/60$ simulations. If we compare
    BIC with an oracle learner which always learns the correct number of arcs,
    we find that BIC learns networks that are too sparse ($p$-value: $3.7\e{-9}$).
    Only for $\sfrac{n}{p} = 5$ the underfitting stops being significant
    ($p$-value: $0.09$).

  \item As we argued in Section \ref{sec:bds}, replacing U with MU results in 
    DAGs with fewer arcs for all $60/60$ simulations. Replacing BDeu with BDs
    results in fewer arcs in $32/60$ simulations for $\alpha = 1$ and in $59/60$
    for $\alpha = 10$, which suggests that the overfitting observed for U+BDeu
    can be attributed to both U and BDeu.

\end{itemize}
The rescaled predictive log-likelihoods in Table \ref{tab:loglik} show that 
U+BDeu never outperforms MU+BDs for $\sfrac{n}{p} < 1.0$ for the same $\alpha$;
for larger $\sfrac{n}{p}$ all scores are tied, and are not reported for brevity.
Even so, the difference between MU+BDs and U+BDeu is too small to be significant
($p$-value: $0.040$ for $\alpha = 1$, $0.0027$ for $\alpha = 10$). The same is
true for BIC ($p$-value: $0.15$; 22/30). U+BDeu for $\alpha_s$ is at best tied
with the corresponding score for  $\alpha = 1$ or $\alpha = 10$. The overall 
best score is MU+BDs for $7/10$ BNs and BIC for the remaining $3/10$.

\subsection{Comparing Different Sparsity Levels for MU}
\label{sec:sim-mu}

We now discuss the effect of the choice of the $\beta$ parameter of MU on both
BDeu and BDs. The results for BDeu are reported in Tables \ref{tab:shd2} (SHD),
\ref{tab:narcs2} (number of arcs) and \ref{tab:loglik2} (predictive 
log-likelihood). We observe the following:
\begin{itemize}
  \item MU with $\beta = \fh$ does not significantly outperform any of $c = 1,
    2, 5$ in terms of SHD for $\alpha = 1$ ($p$-value: $0.99$, $0.66$, $0.99$;
    12/60, 34/60, 18/60) or $\alpha = 10$ ($p$-value: $0.99$, $0.99$, $0.99$;
    7/60, 8/60, 17/60). On the other hand, $c = 1$ significantly outperforms
    $\beta = \fh$ ($p$-value: $2.5\e{-4}$; 53/60). This suggests that enforcing
    sparsity with $c = 1$ is beneficial and results in more accurate structure
    learning than the median-probability prior.

  \item MU with $c = 1, 5$ outperform $c = 2$ in terms of SHD for $\alpha = 1$
    ($p$-value: $3.6\e{-8}$, $1.3\e{-7}$; 50/60, 47/60), but for $\alpha = 10$,
    $c = 1$ outperforms $c = 2$ ($p$-value: $5.4\e{-8}$; 51/60) which in turn
    outperforms $c = 5$ ($p$-value: $2.2\e{-8}$; 54/60). This suggests that 
    enforcing increasing levels of sparsity in the prior does not monotonically
    reduce SHD for $\alpha = 1$, but it does that for $\alpha = 10$. 

    It does, however, reduce the number of arcs as expected. For $\alpha = 1$,
    $c = 1$ produces sparser DAGs than $c = 2$ ($p$-value: $8.3\e{-8}$; 49/60),
    $c = 2$ produces sparser DAGs than $c = 5$ ($p$-value: $3.1\e{-5}$; 48/60),
    and $c = 5$ produces sparser DAGs than $\beta = \fh$ ($p$-value: $8.7\e{-5}$;
    43/60). The situation is similar for $\alpha = 10$
    ($p$-value: $3.8\e{-10}$, $4.2\e{-7}$, $1.4\e{-4}$; 52/60, 54/60, 45/60).

  \item There is no correspondence between the number of arcs expected in the
    prior and the number of arcs present in the learned DAGs, which is roughly
    the same for all $c = 1, 2, 5$. Even $\beta = \fh$ results in only $1.26$
    as many arcs as $c = 1$. Furthermore, there is no apparent relationship
    between the value of $c$ that gives the best SHD and $|\AR|$: for instance,
    WATER has $|\AR|/N = 2.06$ but $c = 1$ results in lower SHDs than $c = 2$,
    while ANDES has $|\AR|/N = 1.51$ but $c = 5$ has the lowest SHD.

  \item We can still recommend the use $\alpha = 1$ over $\alpha = 10$ for BDeu
    when using MU since the former produces significantly smaller SHDs than the
    latter for $\beta = \fh$ ($p$-value: $6.7\e{-6}$; 53/60), $c = 1$ ($p$-value:
    $9.8\e{-7}$; 43/60), $c = 5$ ($p$-value: $3.8\e{-10}$; 51/60), but not for
    $c = 2$ ($p$-value: $0.89$; 0/60) since almost all SHDs are tied in that
    case. 

  \item Even though SHD improves when moving away from the median-probability
    prior, additional sparsity does not improve predictive log-likelihood for
    either $\alpha = 1$ or $\alpha = 10$ ($p$-values: $0.06$, $0.07$, $0.09$,
    for $c = 1, 2, 5$ for both).

\end{itemize}
The results for BDs are reported in Tables \ref{tab:shd3} (SHD), \ref{tab:narcs3}
(number of arcs) and \ref{tab:loglik3} (predictive log-likelihood), and lead to
similar considerations as the above.
\begin{itemize}
  \item All of $c = 1, 2, 5$ produce significantly lower SHD values than 
    $\beta = \fh$ for $\alpha = 1$ ($p$-values: $2.0\e{-4}$, $2.3\e{-4}$, 
    $3.9\e{-4}$; 46/60, 45/60, 40/60) and for $\alpha = 10$ ($p$-values:
    $1.2\e{-3}$, $1.6\e{-3}$, $2.9\e{-3}$; 52/60, 51/60, 45/60). For $\alpha = 1$,
    SHD for $c = 1$ is not significantly better than for $c = 2$ ($p$-value:
    $0.71$; 28/60), but it is for $c = 2$ compared to $c = 5$ ($p$-value:
    $8.4\e{-4}$; 38/60). For $\alpha = 10$, SHD for $c = 1$ is better than for
    $c = 2$ ($p$-value: $3.1\e{-7}$; 49/60) which is better than $c = 5$
    ($p$-value: $3.6\e{-7}$; 50/60). This again suggests enforcing sparsity
    improves the accuracy of BN structure learning.

  \item For BDs, $\alpha = 1$ does not significantly outperform $\alpha = 10$ in 
    terms of SHD for $\beta = \fh$ ($p$-value: $0.012$, 42/60) nor for any
    of $c = 1, 2, 5$ ($p$-values: $0.85$, $0.49$, $0.049$; 30/60, 36/60, 40/60).
    In all cases we $\alpha = 1$ produces smaller SHDs than $\alpha = 10$ in 
    more than half of the simulations, but the difference is small enough that
    it does not reach statistical significance.

  \item As was the case for BDeu, there is no correspondence between the number
    of arcs expected in the prior and the number of arcs present in the learned
    DAGs, and there is no apparent relationship between the value of $c$ that
    gives the best SHD and $|\AR|$.

  \item There is no improvement in the predictive log-likelihood for any of
    $c = 1, 2, 5$ over $\beta = \fh$ for either $\alpha = 1$ or $\alpha = 0.10$
    (all $p$-values are greater than $0.99$).
 
\end{itemize}

\section{Conclusions and Discussion}

In this paper we proposed a new posterior score for discrete BN structure 
learning. We defined it as the combination of a new prior over the space of
DAGs, the ``marginal uniform'' (MU) prior, and of a new empirical Bayes 
marginal likelihood, which we call ``Bayesian Dirichlet sparse'' (BDs). Both
have been designed to address the inconsistent behaviour of the classic 
uniform (U) prior and of BDeu, without requiring any prior information on the
probabilistic structure of the data. 

Issues arising from the use of BDeu have been explored by \citet{silander},
\citet{jaakkola}, \citet{ueno} and \citet{suzuki16} among others. In particular,
our aim was to prevent the inclusion of spurious arcs, particularly in the 
common case of small, sparse data and small imaginary sample sizes. We
complemented the results presented in the papers above, and in particular in
\citet{suzuki16}, by investigating how BDeu model selection is inconsistent
and can assign different scores to DAGs that encode exactly the same information
from the data. From an information theoretic perspective, we find that the
reason of this inconsistency is that the prior distribution assumed in BDeu
lead to a biased estimation of the entropy of BNs learned from sparse data.
From a more probabilistic perspective, we can equivalently say that in such
cases part of the prior probability mass is lost for models with unobserved
parent configurations for at least one node, and this makes Bayesian model
selection inconsistent. These new results motivated the construction of BDs
and the assumption of a piecewise uniform prior on the parameter space.

We also used results from our previous work \citep{ba12} on the properties of U
to motivate the introduction of MU. Firstly, score-based structure learning
explored the space of DAGs with single-arc moves, and this leads to a probability
of inclusion $\pra + \pla = \sfrac{2}{3}$ for each arc and in turn to dense DAGs.
Secondly, arcs incident on a common node are correlated in U, which means that
the inclusion of each false positive leads to an increased probability of more
false positive, making errors in structure learning cascade and multiply.
To address these problems we propose MU and assume independent priors for each
arc with $\pra + \pla \leqslant \fh$. In this context we investigate both the
median-probability prior and sparsity inducing priors with $O(N)$ expected arcs.

In an extensive simulation study based on $10$ reference BNs we find that MU+BDs
(assuming $\beta = \fh$) outperforms U+BDeu for all combinations of BN and sample
sizes in the quality of the learned DAGs. Predictive accuracy is not significantly
different between the two scores. We confirm that $\alpha = 1$ is a good default
as suggested by \citet{ueno}, for both BDeu and BDs. As for MU, the
median-probability prior with $\beta = \fh$ is outperformed in terms of SHD by
the $\beta$ corresponding to $c = 1$. Clearly, this result only generalises to
data whose underlying DAGs are sparse, as is the case for the reference BNs.

This is achieved without increasing the computational complexity of the
posterior score, since MU+BDs can be computed in the same time as U+BDeu. In
this respect, the posterior score we propose is preferable to similar proposals
in the literature. For instance, the NIP-BIC score from \citet{ueno2} and the
NIP-BDe/Expected log-BDe scores from \citet{ueno3} outperform BDeu but at a
significant computational cost. The same is true for the optimal $\alpha$
proposed by \citet{alphastar} for BDeu, whose estimation requires multiple runs
of the structure learning algorithm to converge. The Max-BDe and Min-BDe scores
in \citet{minbde} overcome in part the limitations of BDeu by optimising for
either goodness of fit at the expense of predictive accuracy, or vice versa.
As a further term of comparison, we also included BIC in the simulation; 
while it outperforms U+BDeu in some circumstances and it is computationally
efficient, MU+BDs is better overall in the DAGs it learns and competitive in
predictive accuracy.

\acks{We would like to acknowledge the anonymous referees and the attendees
of the International Conference on Probabilistic Graphical Models (PGM) for
their observations and the stimulating discussions on the topic of this
paper.}

\newpage
\appendix

\linespread{1}

\section{Proofs}
\label{app:proofs}

\begin{proof}{\bf of Theorem \ref{thm:h-monotonic}.}
The posterior probabilities $p_{ijk}^{(\alpha)}$ can be rewritten as
\begin{equation*}
  \pijk^{(\asi)} = \frac{\asi + n_{ijk}}{r_i \asi + n_{ij}}
    = \frac{\asi}{r_i \asi + n_{ij}} \cdot \frac{\asi}{r_i \asi} + 
      \frac{n_{ij}}{r_i \asi + n_{ij}} \cdot \frac{n_{ijk}}{n_{ij}}
    = \lambda \frac{1}{r_i} + (1 - \lambda_{ij}) p_{ijk}
\end{equation*}
with
\begin{equation*}
  \lambda_{ij} = \frac{r_i \asi}{r_i \asi + n_{ij}},
\end{equation*}
which is a weighted average between the uniform prior ($\sfrac{1}{r_i}$) and
the observed empirical frequencies ($p_{ijk}$). We can decompose the posterior
entropy along the same lines, and use the concavity of entropy \citep{itheory}:
\begin{multline*}
  \H(\XPi; \alpha) 
    = \sum_{j = 1}^{q_i} \H(\XPi = j; \alpha) 
    = \sum_{j = 1}^{q_i} \H\left( (1 - \lambda_{ij})(\XPi = j) + 
                                   \lambda_{ij} U \right) \geqslant \\
    \geqslant \sum_{j = 1}^{q_i} (1 - \lambda_{ij}) \H(\XPi = j) +
                                  \lambda_{ij} \H(U).
\end{multline*}
First, we note that this implies $\H(\XPi; \alpha) \geqslant \H(\XPi)$ for any
$\alpha > 0$:
\begin{multline*}
  \sum_{j = 1}^{q_i} (1 - \lambda_{ij}) \H(\XPi = j) + \lambda_{ij} \H(U) \geqslant \\
  \geqslant \sum_{j = 1}^{q_i} (1 - \lambda_{ij}) \H(\XPi = j) +
                                \lambda_{ij} \H(\XPi = j) \\
  = \sum_{j = 1}^{q_i} \H(\XPi = j) = \H(\XPi),
\end{multline*}
with the equality holding iff $\H(U) = \H(\XPi = j)$. Furthermore, we can compose
weighted averages and write
\begin{align*}
  p_{ijk}^{(\beta)} 
    = \delta_{ij} \frac{1}{r_i} + (1 - \delta_{ij}) \pijk^{(\alpha)}
    = \delta_{ij}\lambda_{ij} \frac{1}{r_i} + (1 - \delta_{ij}\lambda_{ij}) \pijk
\end{align*}
and since $\beta > \alpha > 0$ we have
\begin{equation*}
  \delta_{ij}\lambda_{ij} = \frac{\beta_{ij}}{\beta_{ij} + n_{ij}} > 
    \frac{r_i \asi}{r_i \asi + n_{ij}} > 0.
\end{equation*}
Proceeding along the same lines as above we obtain
\begin{equation*}
  \H(\XPi; \beta)
    \geqslant \sum_{j = 1}^{q_i} (1 - \delta_{ij}) \H(\XPi = j; \alpha) + \delta_{ij} \H(U)
    \geqslant \H(\XPi; \alpha)
\end{equation*}
with equality iff $\H(U) = \H(\XPi = j; \alpha)$ as required.
\end{proof}

\begin{proof}{\bf of Theorem \ref{thm:bdeu-monotonic}.}
A Laurent series expansion for $\bsi = \beta / (r_i q_i)$, $\bsi \to 0$ for 
fixed $r_i$ gives
\begin{align}
  &\frac{\Gamma(r_i \bsi)}{\Gamma(r_i \bsi + n_{ij})} 
     \approx \frac{1}{r_i \bsi \Gamma(n_{ij})}&
  &\text{and}&
  &\frac{\Gamma(\bsi + n_{ijk})}{\Gamma(\bsi)}
     \approx \bsi \Gamma(n_{ijk});
\label{eq:laurent}
\end{align}
Since $\beta > \alpha \to 0$, $\bsi > \asi \to 0$ we have
\begin{align*}
  \prod_{j : n_{ij} > 0} \left[ 
    \frac{1}{r_i \bsi \Gamma(n_{ij})} \prod_{k: n_{ijk} > 0} \bsi \Gamma(n_{ijk})
  \right] &>
  \prod_{j : n_{ij} > 0} \left[ 
    \frac{1}{r_i \asi \Gamma(n_{ij})} \prod_{k: n_{ijk} > 0} \asi \Gamma(n_{ijk})
  \right] \\
  \prod_{j : n_{ij} > 0} \left[ 
    \frac{1}{\bsi} \prod_{k: n_{ijk} > 0} \bsi
  \right] &>
  \prod_{j : n_{ij} > 0} \left[ 
    \frac{1}{\asi} \prod_{k: n_{ijk} > 0} \asi
  \right] \\
  (\bsi)^{-\tq}(\bsi)^{\sum_k \tilde{r}_{ij}} &>
  (\asi)^{-\tq}(\asi)^{\sum_k \tilde{r}_{ij}} \\
  (\bsi)^{\dEP{X_i, \G}} &> (\asi)^{\dEP{X_i, \G}}
\end{align*}
as required. If $\dEP{X_i, \G} = \sum_k \tilde{r}_{ij} - \tq = 0$,
there is only a single $n_{ijk} > 0$ for each $n_{ij} > 0$ so
$n_{ijk} = n_{ij}$. Then
\begin{equation*}
  \prod_{j : n_{ij} > 0} \left[ 
    \frac{1}{r_i \asi \Gamma(n_{ij})}
    \prod_{k: n_{ijk} > 0} \asi \Gamma(n_{ijk})
  \right] =
  \prod_{j : n_{ij} > 0} \left[ 
    \frac{1}{\cancel{r_i \asi \Gamma(n_{ij})}}
    \frac{\cancel{r_i \asi \Gamma(n_{ij})}}{r_i}
  \right] =
  \left(\frac{1}{r_i}\right)^{\tq}
\end{equation*}
which proves the second case.
\end{proof}

\begin{proof}{\bf of Theorem \ref{thm:bdlink}.}
Substituting the approximation from \mref{eq:laurent} in \mref{eq:bds} yields
\begin{multline*}
  \BDs(\XPi; \alpha) \approx \prod_{j : n_{ij} > 0}
    \left[ 
      \frac{1}{r_i \tai \Gamma(n_{ij})}
      \prod_{k: n_{ijk} > 0} \tai\Gamma(n_{ijk})
    \right] = \\ = \prod_{j : n_{ij} > 0}
    \left[ 
      \frac{1}{r_i \asi \Gamma(n_{ij})}
      \prod_{k: n_{ijk} > 0} \asi \Gamma(n_{ijk})
    \right] \cdot \left( \frac{\tq}{q_i} \right)^{\tq} \left( \frac{q_i}{\tq} \right)^{\sum_j \tilde{r}_{ij}} = \\ =
    \BDeu(\XPi; \alpha) \cdot \left( \frac{q_i}{\tq} \right)^{\sum_j \tilde{r}_{ij} - \tq } =
    \BDeu(\XPi; \alpha) \cdot \left( \frac{q_i}{\tq} \right)^{\dEP{X_i, \G}}
\end{multline*}
using the fact that $\tai = \asi (q_i / \tq)$.
\end{proof}

\begin{proof}{\bf of Theorem \ref{thm:asymeq}.}
Under condition \ref{thm:eq:one}, all $n_{ijk} > 0$ by the law of large numbers.
Therefore all $n_{ij} > 0$ and $\tq = q_i$, leading to
$\BDs(\XPi; \alpha) = \BDeu(\XPi; \alpha)$. As for condition
\ref{thm:eq:two}, $\alpha \to \infty$ implies $r_i \tai \to \infty$ and 
$\tai \to \infty$ since $\tai = \alpha / (r_i \tq)$ and $r_i, q_i$ are fixed for
a given network. Then by Stirling's approximation we have that
\begin{align*}
  &\frac{\Gamma(r_i \tai)}
        {\Gamma(r_i \tai + n_{ij})} \approx (r_i \tai)^{-n_{ij}}&
  &\text{and}&
  &\frac{\Gamma(\tai + n_{ijk})}
        {\Gamma(\tai)} \approx (\tai)^{n_{ijk}}
\end{align*}
and by substituting the above in \mref{eq:bds} we obtain
\begin{multline*}
  \BDs(\XPi; \alpha) \approx
    \prod_{j:n_{ij} > 0} \left[
      (r_i \tai)^{-n_{ij}} \prod_{k:n_{ijk} > 0} (\tai)^{n_{ijk}}
    \right] = \\ =
    \prod_{j:n_{ij} > 0} \left[
      (r_i \asi)^{-n_{ij}} \prod_{k:n_{ijk} > 0} (\asi)^{n_{ijk}}
      \right]
      \cancel{\left[ 
        \left(\frac{q_i}{\tq}\right)^{-n_{ij}} \prod_{k:n_{ijk} > 0}  \left(\frac{q_i}{\tq} \right)^{n_{ijk}}
      \right]} \approx \\ \approx
    \prod_{j:n_{ij} > 0} \left[
      \frac{\Gamma(r_i \asi)}{\Gamma(r_i \asi + n_{ij})}
      \prod_{k:n_{ijk} > 0} \frac{\Gamma(\asi + n_{ijk})}{\Gamma(\asi)}
    \right]
   = \BDeu(\XPi; \alpha).
\end{multline*}
\end{proof}

\begin{proof}{\bf of Theorem \ref{thm:threshold}.}
The Bayes factor for BDs can be obtained by combining \mref{eq:convergence},
\mref{eq:bdeu-ratio} and \mref{eq:relationship}:
\begin{equation*}
  \frac{\BDs(\XPi \cup X_l; \alpha)}{\BDs(\XPi; \alpha)} =
  \frac{(q_i / \tq)^{\dEP{X_i, \Gp}} \alpha^{\dEP{\Gp}}}
       {(q'_i / \tilde{q}'_i)^{\dEP{X_i, \Gm}} \alpha^{\dEP{\Gm}}} =
  \frac{\alpha^{\dEP{\Gp} + \dEP{X_i, \Gp} \log_\alpha q_i / \tq}}
       {\alpha^{\dEP{\Gm} + \dEP{X_i, \Gm} \log_\alpha q'_i / \tilde{q}'_i}}.
\end{equation*}
Following \mref{eq:bdeu-ratio} we can then write
\begin{equation*}
  \frac{\BDs(\XPi \cup X_l; \alpha)}{\BDs(\XPi; \alpha)}
  \to
  \left\{
  \begin{aligned}
  0 & & \text{if $d_\mathrm{EDF} > 0$}\\
  +\infty & & \text{if $d_\mathrm{EDF} < 0$} 
  \end{aligned}
  \right.
\end{equation*}
since we operate under the same assumptions and since the left-hand term differs
from that in \mref{eq:bdeu-ratio} by a constant multiplier that does not depend
on $\alpha$. The Bayes factor then diverges if and only if
\begin{equation*}
  \dEP{\Gp} + \dEP{X_i, \Gp} \log_\alpha q_i / \tq  < 
    \dEP{\Gm} + \dEP{X_i, \Gm} \log_\alpha q'_i / \tilde{q}'_i 
\end{equation*}
which is equivalent to
\begin{equation*}
  \underbrace{\dEP{\Gp} - \dEP{\Gm}}_{d_\mathrm{EDF}} < 
  \underbrace{\dEP{X_i, \Gm} \log_\alpha q'_i / \tilde{q}'_i - 
              \dEP{X_i, \Gp} \log_\alpha q_i / \tq
  }_{-\log_\alpha (\Prob(\Gp) / \Prob(\Gm))}
\end{equation*}
and it converges to zero otherwise.
\end{proof}

\clearpage
\section{Additional Simulation Results}
\label{app:tables}

In this appendix we report the detailed results of the simulations described in
Section \ref{sec:sim}, organised in tables as follows.

\begin{itemize}
  \item Table \ref{tab:shd}: average SHD from $\GR$, discussed in Section 
    \ref{sec:sim-bds}. This table compares U+BDeu, U+BDs, MU+BDeu, MU+BDs and
    BIC. For MU, $\beta$ is fixed to $\fh$ to give the median-probability model.
  \item Table \ref{tab:narcs}: average number of arcs relative to $\GR$,
    discussed in Section \ref{sec:sim-bds}. This table compares U+BDeu, U+BDs,
    MU+BDeu, MU+BDs and BIC like Table \ref{tab:shd}.
  \item Table \ref{tab:loglik}: average predictive log-likelihood values from
    Section \ref{sec:sim-bds}. This table compares U+BDeu, U+BDs, MU+BDeu,
    MU+BDs and BIC like Tables \ref{tab:shd} and \ref{tab:narcs}.
  \item Table \ref{tab:shd2}: average SHD for BDeu and different MU priors, 
    discussed in Section \ref{sec:sim-mu}. Results for U from Table
    \ref{tab:shd} are included as well for ease of reference.
  \item Table \ref{tab:shd3}: average SHD for BDs and different MU priors,
    discussed in Section \ref{sec:sim-mu}. Results for U from Table
    \ref{tab:narcs} are included as well for ease of reference.
  \item Table \ref{tab:narcs2}: average number of arcs for BDeu and different
    MU priors, discussed in Section \ref{sec:sim-mu}. Results for U from Table
    \ref{tab:narcs} are included as well for ease of reference.
  \item Table \ref{tab:narcs3}: average number of arcs for BDs and different
    MU priors, discussed in Section \ref{sec:sim-mu}. Results for U from Table
    \ref{tab:narcs} are included as well for ease of reference.
  \item Table \ref{tab:loglik2}: average predictive log-likelihood values for
    BDeu and different MU priors, discussed in Section \ref{sec:sim-mu}.
    Results for U from Table \ref{tab:loglik} are included as well for ease of
    reference.
  \item Table \ref{tab:loglik3}: average predictive log-likelihood values for
    BDs and different MU priors, discussed in Section \ref{sec:sim-mu}. Results
    for U from Table \ref{tab:loglik} are included as well for ease of reference.
\end{itemize}

\begin{table}[p]
\begin{center}
  \scriptsize
\begin{tabular}{lr|r|rrr|rr|rr|rr}

    \hline
    \multirow{2}{*}{NETWORK} & \multirow{2}{*}{$n/p$} & \multicolumn{1}{|c|}{BIC} & \multicolumn{3}{|c|}{U + BDeu} & \multicolumn{2}{|c|}{U + BDs} & \multicolumn{2}{|c|}{MU + BDeu} & \multicolumn{2}{|c}{MU + BDs} \\
                             &                        &                           & 1 & $\alpha_S$ & 10            & 1 & 10                         & 1 & 10                          & 1 & 10                         \\ 
    \hline
    \multirow{7}{*}{ALARM} 
            &  0.1 &  55.5 &   78.0 &   80.5 &   112.7 &     64.2 &     87.3 & \bf 53.0 &      83.5 &  \bf 53.0 &       65.5 \\
            &  0.2 &  50.8 &   49.2 &   56.1 &    92.8 &     49.5 &     75.2 & \bf 39.6 &      68.3 &  \bf 39.6 &       56.2 \\
            &  0.5 &  40.8 &   35.5 &   41.9 &    72.0 &     34.9 &     61.5 & \bf 31.3 &      53.5 &  \bf 31.3 &       46.1 \\
            &  1.0 &  33.7 &   31.9 &   37.6 &    62.6 &     29.1 &     51.8 & \bf 27.1 &      49.8 &  \bf 27.1 &       42.1 \\
            &  2.0 &  28.1 &   26.3 &   31.9 &    53.1 &     23.1 &     44.5 & \bf 22.9 &      41.0 &  \bf 22.9 &       36.5 \\
            &  5.0 &  22.6 &   24.4 &   30.1 &    41.6 &     20.9 &     35.0 & \bf 20.4 &      31.6 &  \bf 20.4 &       28.9 \\
\hline
  \end{tabular}
  \caption{Average SHD from $\GR$ (lower is better, best in bold).}
  \label{tab:shd}
\end{center}
\end{table}

\begin{table}[p]
\begin{center}
  \scriptsize
  \begin{tabular}{lr|r|rrr|rr|rr|rr}

    \hline
    \multirow{2}{*}{NETWORK} & \multirow{2}{*}{$n/p$} & \multicolumn{1}{|c|}{BIC} & \multicolumn{3}{|c|}{U + BDeu} & \multicolumn{2}{|c|}{U + BDs} & \multicolumn{2}{|c|}{MU + BDeu} & \multicolumn{2}{|c}{MU + BDs} \\
                             &                        &                           & 1 & $\alpha_S$ & 10            & 1 & 10                         & 1 & 10                          & 1 & 10                         \\
    \hline
    \multirow{7}{*}{ALARM}
            & 0.1 &  0.596 &   1.635 &   1.697 &    2.550 &    1.329 &     1.875 & \bf 1.040 &      1.854 &  \bf 1.040 &       1.351 \\
            & 0.2 &  0.662 &   1.272 &   1.448 &    2.278 &    1.321 &     1.874 & \bf 1.049 &      1.730 &  \bf 1.049 &       1.436 \\
            & 0.5 &  0.746 &   1.174 &   1.290 &    1.993 &    1.213 &     1.775 & \bf 1.060 &      1.605 &  \bf 1.060 &       1.436 \\
            & 1.0 &  0.859 &   1.165 &   1.302 &    1.830 &    1.180 &     1.667 & \bf 1.071 &      1.553 &  \bf 1.071 &       1.426 \\
            & 2.0 &  0.972 &   1.117 &   1.236 &    1.664 &    1.098 &     1.528 & \bf 1.064 &      1.445 &  \bf 1.064 &       1.377 \\
            & 5.0 &  1.092 &   1.098 &   1.208 &    1.457 &    1.086 &     1.386 & \bf 1.061 &      1.286 &  \bf 1.061 &       1.252 \\
\hline
  \end{tabular}
  \caption{Average number of arcs (rescaled by $|\AR|$; closer to $1$ is better, best in bold).}
  \label{tab:narcs}
\end{center}
\end{table}

\begin{table}[p]
\begin{center}
  \scriptsize
  \begin{tabular}{lr|r|rrr|rr|rr|rr}

    \hline
    \multirow{2}{*}{NETWORK} & \multirow{2}{*}{$n/p$} & \multicolumn{1}{|c|}{BIC} & \multicolumn{3}{|c|}{U + BDeu} & \multicolumn{2}{|c|}{U + BDs} & \multicolumn{2}{|c|}{MU + BDeu} & \multicolumn{2}{|c}{MU + BDs} \\
                             &                        &                           & 1 & $\alpha_S$ & 10            & 1 & 10                         & 1 & 10                          & 1 & 10                         \\ 
    \hline
    \multirow{3}{*}{ALARM}
             & 0.1 &    1.54 &     1.67 &     1.68 &      1.85 &      1.67 &       1.80 &   \bf 1.51 &        1.69 &    \bf 1.51 &         1.60 \\
             & 0.2 &    1.33 &     1.32 &     1.34 &      1.44 &      1.35 &       1.43 &   \bf 1.29 &        1.36 &    \bf 1.29 &         1.34 \\
             & 0.5 &    1.21 &     1.17 &     1.17 &      1.21 &      1.17 &       1.20 &   \bf 1.16 &        1.18 &    \bf 1.16 &         1.17 \\
    \hline
    \multirow{3}{*}{ANDES}
             & 0.1 & \bf 11.12 &    13.14 &    17.56 &     18.59 &     14.75 &      24.40 &      11.90 &       15.77 &       11.90 &        17.77 \\
             & 0.2 & \bf 10.00 &    10.56 &    11.53 &     11.96 &     10.88 &      13.30 &      10.16 &       11.13 &       10.16 &        11.47 \\
             & 0.5 & \bf  9.50 &     9.60 &     9.80 &      9.96 &      9.63 &      10.07 &       9.53 &        9.73 &        9.53 &         9.74 \\
\hline
  \end{tabular}
  \caption{Average predictive log-likelihood (rescaled by $-10000$; lower is
    better, best in bold). $\sfrac{n}{p} = 1.0, 2.0, 5.0$ showed the same 
    value for all scores and are omitted for brevity.}
  \label{tab:loglik}
\end{center}
\end{table}

\begin{table}[p]
\begin{center}
  \scriptsize
  \begin{tabular}{lr|rr|rr|rr|rr|rr}

    \hline
    \multirow{2}{*}{NETWORK} & \multirow{2}{*}{$n/p$} & \multicolumn{2}{|c|}{U} & \multicolumn{2}{|c|}{MU ($\beta = \fh$)} & \multicolumn{2}{|c}{MU ($c = 1$)} & \multicolumn{2}{|c}{MU ($c = 2$)} & \multicolumn{2}{|c}{MU ($c = 5$)} \\
                             &                        & 1 & 10                  & 1 & 10                                   & 1 & 10                            & 1 & 10                            & 1 & 10 \\ 
    \hline
    \multirow{7}{*}{ALARM} 
            & 0.1 &  78.0 &  112.7 &     53.0 &  83.5 &  \bf 42.9 &      47.1 &      53.1 &     53.1 &      46.8 & 68.8 \\
            & 0.2 &  49.2 &   92.8 &     39.6 &  68.3 &  \bf 33.8 &      40.7 &      46.3 &     46.3 &      36.6 & 59.3 \\
            & 0.5 &  35.5 &   72.0 &     31.3 &  53.5 &  \bf 27.2 &      33.1 &      37.9 &     37.9 &      29.2 & 48.2 \\
            & 1.0 &  31.9 &   62.6 &     27.1 &  49.8 &  \bf 25.6 &      32.1 &      35.8 &     35.8 &      27.1 & 45.2 \\
            & 2.0 &  26.3 &   53.1 &     22.9 &  41.0 &  \bf 22.5 &      28.9 &      31.7 &     31.7 &      23.6 & 37.0 \\
            & 5.0 &  24.4 &   41.6 & \bf 20.4 &  31.6 &      21.9 &      26.9 &      28.1 &     28.1 &      22.3 & 31.3 \\
\hline
  \end{tabular}
  \caption{Average SHD for BDeu and different MU priors (lower is better, best in bold).}
  \label{tab:shd2}
\end{center}
\end{table}

\begin{table}[p]
\begin{center}
  \scriptsize
  \begin{tabular}{lr|rr|rr|rr|rr|rr}

    \hline
    \multirow{2}{*}{NETWORK} & \multirow{2}{*}{$n/p$} & \multicolumn{2}{|c|}{U} & \multicolumn{2}{|c|}{MU ($\beta = \fh$)} & \multicolumn{2}{|c}{MU ($c = 1$)} & \multicolumn{2}{|c}{MU ($c = 2)$} & \multicolumn{2}{|c}{MU ($c = 5$)} \\
                             &                        & 1 & 10                  & 1 & 10                                   & 1 & 10                            & 1 & 10                            & 1 & 10 \\ 
    \hline
    \multirow{7}{*}{ALARM} 
            &  0.1 &      64.2 &     87.3 &      53.0 &      65.5 &  \bf 43.6 &     45.6 &      44.2 &      48.8 &      48.9 &  56.6 \\
            &  0.2 &      49.5 &     75.2 &      39.6 &      56.2 &  \bf 34.2 &     37.3 &  \bf 34.2 &      40.0 &      37.2 &  47.4 \\
            &  0.5 &      34.9 &     61.5 &      31.3 &      46.1 &  \bf 24.5 &     29.7 &      24.8 &      33.0 &      26.8 &  40.3 \\
            &  1.0 &      29.1 &     51.8 &      27.1 &      42.1 &  \bf 21.8 &     25.7 &      21.9 &      28.8 &      23.4 &  36.7 \\
            &  2.0 &      23.1 &     44.5 &      22.9 &      36.5 &  \bf 16.6 &     23.3 &  \bf 16.6 &      25.7 &      17.1 &  31.2 \\
            &  5.0 &      20.9 &     35.0 &      20.4 &      28.9 &  \bf 18.5 &     21.5 &      18.7 &      22.5 &      18.9 &  25.6 \\
\hline
  \end{tabular}
  \caption{Average SHD for BDs and different MU priors (lower is better, best in bold).}
  \label{tab:shd3}
\end{center}
\end{table}

\begin{table}[p]
\begin{center}
  \scriptsize
  \begin{tabular}{lr|rr|rr|rr|rr|rr}

    \hline
    \multirow{2}{*}{NETWORK} & \multirow{2}{*}{$n/p$} & \multicolumn{2}{|c|}{U} & \multicolumn{2}{|c|}{MU ($\beta = \fh$)} & \multicolumn{2}{|c}{MU ($c = 1$)} & \multicolumn{2}{|c}{MU ($c = 2$)} & \multicolumn{2}{|c}{MU ($c = 5$)} \\
                             &                        & 1 & 10                  & 1 & 10                                   & 1 & 10                            & 1 & 10                            & 1 & 10 \\ 
    \hline
    \multirow{7}{*}{ALARM}
            & 0.1 &     1.635 &     2.550 & \bf 1.040 &     1.854 &     0.638 &     0.876 & 0.698 &     1.073 &     0.841 & 1.497 \\
            & 0.2 &     1.272 &     2.278 & \bf 1.038 &     1.730 &     0.766 &     1.049 & 0.821 &     1.201 &     0.947 & 1.510 \\
            & 0.5 &     1.174 &     1.993 &     1.060 &     1.605 &     0.876 &     1.121 & 0.911 &     1.235 & \bf 0.979 & 1.463 \\
            & 1.0 &     1.165 &     1.830 &     1.071 &     1.553 & \bf 0.995 &     1.164 & 0.975 &     1.253 &     1.040 & 1.457 \\
            & 2.0 &     1.117 &     1.664 &     1.064 &     1.445 & \bf 1.039 &     1.151 & 1.047 &     1.209 &     1.063 & 1.336 \\
            & 5.0 &     1.098 &     1.457 &     1.061 &     1.286 & \bf 1.053 &     1.129 & 1.055 &     1.158 &     1.061 & 1.226 \\
\hline
  \end{tabular}
  \caption{Average number of arcs for BDeu and different MU priors (closer to $1$ is better, best in bold).}
  \label{tab:narcs2}
\end{center}
\end{table}

\begin{table}[p]
\begin{center}
  \scriptsize
  \begin{tabular}{lr|rr|rr|rr|rr|rr}

    \hline
    \multirow{2}{*}{NETWORK} & \multirow{2}{*}{$n/p$} & \multicolumn{2}{|c|}{U} & \multicolumn{2}{|c|}{MU ($\beta = \fh$)} & \multicolumn{2}{|c}{MU ($c = 1$)} & \multicolumn{2}{|c}{MU ($c = 2$)} & \multicolumn{2}{|c}{MU ($c = 5$)} \\
                             &                        & 1 & 10                  & 1 & 10                                   & 1 & 10                            & 1 & 10                           & 1 & 10 \\ 
    \hline
    \multirow{7}{*}{ALARM}
            & 0.1 &     1.329 &     1.875 & \bf 1.040 &     1.351 &     0.649 &     0.736 & 0.723 &     0.862 & 0.911     & 1.121 \\
            & 0.2 &     1.321 &     1.874 &     1.049 &     1.436 &     0.797 &     0.909 & 0.847 &     1.016 & \bf 0.974 & 1.236 \\
            & 0.5 &     1.213 &     1.775 &     1.060 &     1.436 &     0.885 &     1.053 & 0.911 &     1.140 & \bf 0.988 & 1.302 \\
            & 1.0 &     1.180 &     1.667 &     1.071 &     1.426 &     0.962 &     1.107 & 0.983 &     1.182 & \bf 1.018 & 1.337 \\
            & 2.0 &     1.098 &     1.528 &     1.064 &     1.377 & \bf 1.017 &     1.107 & 1.023 &     1.153 &     1.033 & 1.266 \\
            & 5.0 &     1.086 &     1.386 &     1.061 &     1.252 & \bf 1.045 &     1.096 & 1.048 &     1.117 &     1.051 & 1.182 \\
\hline
  \end{tabular}
  \caption{Average number of arcs for BDs and different MU priors (closer to $1$ is better, best in bold).}
  \label{tab:narcs3}
\end{center}
\end{table}

\begin{table}[p]
\begin{center}
  \scriptsize
  \begin{tabular}{lr|rr|rr|rr|rr|rr}

    \hline
    \multirow{2}{*}{NETWORK} & \multirow{2}{*}{$n/p$} & \multicolumn{2}{|c|}{U} & \multicolumn{2}{|c|}{MU ($\beta = \fh$)} & \multicolumn{2}{|c}{MU ($c = 1$)} & \multicolumn{2}{|c}{MU ($c = 2$)} & \multicolumn{2}{|c}{MU ($c = 5$)} \\
                             &                        & 1 & 10                  & 1 & 10                                   & 1 & 10                            & 1 & 10                            & 1 & 10 \\ 
    \hline
    \multirow{3}{*}{ALARM}
             & 0.1 &      1.67 &      1.85 &      1.51 &      1.69 & \bf 1.42 &      1.44 &     1.43 &     1.48 &     1.47 & 1.58 \\
             & 0.2 &      1.32 &      1.44 &      1.29 &      1.36 & \bf 1.26 &      1.28 & \bf 1.26 &     1.29 &     1.28 & 1.33 \\
             & 0.5 &      1.17 &      1.21 &      1.16 &      1.18 & \bf 1.15 &  \bf 1.15 & \bf 1.15 &     1.16 & \bf 1.15 & 1.17 \\
    \hline
    \multirow{3}{*}{ANDES}
             & 0.1 &     13.14 &     18.59 &     11.90 &     15.77 &    10.26 & \bf 10.19 &    10.24 &    10.22 &    10.26 & 10.39 \\
             & 0.2 &     10.56 &     11.96 &     10.16 &     11.13 &     9.72 &  \bf 9.69 &     9.70 &     9.72 &     9.71 &  9.81 \\
             & 0.5 &      9.60 &      9.96 &      9.53 &      9.73 &     9.47 &  \bf 9.45 &     9.47 & \bf 9.45 &     9.46 &  9.46 \\
\hline
  \end{tabular}
  \caption{Average predictive log-likelihood for BDeu and different MU priors
    (rescaled by $-10000$; lower is better, best in bold).
    $\sfrac{n}{p} = 1.0, 2.0, 5.0$ showed the same value for all scores and are
    omitted for brevity.}
  \label{tab:loglik2}
\end{center}
\end{table}

\begin{table}[p]
\begin{center}
  \scriptsize
  \begin{tabular}{lr|rr|rr|rr|rr|rr}

    \hline
    \multirow{2}{*}{NETWORK} & \multirow{2}{*}{$n/p$} & \multicolumn{2}{|c|}{U} & \multicolumn{2}{|c|}{MU ($\beta = \fh$)} & \multicolumn{2}{|c}{MU ($c = 1$)} & \multicolumn{2}{|c}{MU ($c = 2$)} & \multicolumn{2}{|c}{MU ($c = 5$)} \\
                             &                        & 1 & 10                  & 1 & 10                                   & 1 & 10                            & 1 & 10                            & 1 & 10 \\ 
    \hline
    \multirow{3}{*}{ALARM}
             & 0.1 &      1.67 &      1.80 &      1.51 &      1.60 & \bf 1.42 &   \bf 1.42 &     1.43 &      1.44 &     1.50 & 1.52 \\
             & 0.2 &      1.35 &      1.43 &      1.29 &      1.34 & \bf 1.27 &   \bf 1.27 & \bf 1.27 &      1.28 &     1.29 & 1.31 \\
             & 0.5 &      1.17 &      1.20 &      1.16 &      1.17 & \bf 1.14 &       1.15 & \bf 1.14 &      1.15 &     1.15 & 1.16 \\
    \hline
    \multirow{3}{*}{ANDES}
             & 0.1 &     14.75 &     24.40 &     11.90 &     17.77 &    10.25 &  \bf 10.19 &    10.24 & \bf 10.19 &    10.26 & 10.34 \\
             & 0.2 &     10.88 &     13.30 &     10.16 &     11.47 &     9.71 &  \bf  9.68 &     9.70 &      9.70 &     9.71 &  9.72 \\
             & 0.5 &      9.63 &     10.07 &      9.53 &      9.74 &     9.47 &   \bf 9.45 &     9.47 &  \bf 9.45 &     9.46 &  9.46 \\
\hline
  \end{tabular}
  \caption{Average predictive log-likelihood for BDs and different MU priors
    (rescaled by $-10000$; lower is better, best in bold).
    $\sfrac{n}{p} = 1.0, 2.0, 5.0$ showed the same value for all scores and
    are omitted for brevity.}
  \label{tab:loglik3}
\end{center}
\end{table}

\end{document}